%% file: main.tex
\documentclass{article}

\newif\ifpreprint
\preprinttrue

\usepackage{microtype}
\usepackage{graphicx}
\usepackage{subfigure}
\usepackage{booktabs}
\usepackage{amsmath,amssymb}
\usepackage[table]{xcolor}
\usepackage{multirow}

\usepackage{cuted}
\usepackage{caption}

\usepackage{xr-hyper}
\makeatletter
\newcommand*{\addFileDependency}[1]{\typeout{(#1)}\@addtofilelist{#1}\IfFileExists{#1}{}{\typeout{No file #1.}}}
\makeatother
\newcommand*{\myexternaldocument}[1]{\externaldocument{#1}\addFileDependency{#1.tex}\addFileDependency{#1.aux}}
\ifpreprint\else
\myexternaldocument{appendix}
\fi
\usepackage{hyperref}
\usepackage{xurl}
\usepackage{fontawesome5}

\ifpreprint
\usepackage[accepted]{mlsys2025}
\else
\usepackage{mlsys2025}
\fi
\renewcommand{\algorithmiccomment}[1]{\hfill$\triangleright$~#1}

\mlsystitlerunning{Sub 1-bit KV Cache Quantization for Looped Transformers}

\begin{document}

\twocolumn[
\mlsystitle{LoopVQ: Sub 1-bit KV Cache Quantization for Looped Transformers}

\mlsyssetsymbol{equal}{*}

\ifpreprint
{\centering
\begin{tabular}{c}
\textbf{Minsoo Cheong} \\
Seoul National University \\
\small\texttt{icycle0409@snu.ac.kr}
\end{tabular}
\hspace{3em}
\begin{tabular}{c}
\textbf{Sungjoo Yoo} \\
Seoul National University \\
\small\texttt{sungjoo@gmail.com}
\end{tabular}
\par}
\hypersetup{pdfauthor={Minsoo Cheong, Sungjoo Yoo}}
\fi

\mlsyscorrespondingauthor{Sungjoo Yoo}{sungjoo@gmail.com}

\mlsyskeywords{Machine Learning, MLSys}

\vskip 0.3in

\begin{abstract}
Looped transformers are a parameter-efficient alternative to standard transformers for language
modeling. By reapplying the same layers across multiple loop steps, recent looped language models match the performance of non-looped models several times their size. While this architecture effectively reduces parameter memory, each loop step still stores its
own keys and values, so the key-value (KV) cache grows linearly with the number of loop steps and
becomes the main memory bottleneck. We observe that the KV cache of looped models is highly redundant along the
loop axis and that the loop steps differ substantially in their sensitivity to quantization. Based
on these observations, we propose \textbf{LoopVQ}, a KV cache vector quantization method for
looped transformers that codes channels of multiple loop steps jointly, and fine-tunes the codebooks to match the output of the full-precision model. At 0.5 bits per element, LoopVQ outperforms existing KV cache vector quantization
methods for standard LLMs on general, long chain-of-thought reasoning, and long-context retrieval
tasks with Ouro and Huginn models. Its design is orthogonal to these methods, and combined with
TaSQ, it retains much of the BF16 accuracy even at 0.25 bits per element. We also implement LoopVQ in SGLang,
where it nearly halves the decode latency at a 32K-token context and raises the peak serving
throughput by up to $4.3\times$ and $6.1\times$ on the 1.4B and 2.6B Ouro models compared to
the BF16 baseline.
\end{abstract}

{\centering
\ifpreprint
\href{https://github.com/mscheong01/LoopVQ}{
  \textcolor{blue}{
    \faGithub\ 
    \texttt{\textbf{Code}: https://github.com/mscheong01/LoopVQ}
  }
}\par
\fi}
]

\input{Sections/1_Introduction}
\input{Sections/2_Preliminaries}
\input{Sections/3_Motivation}
\input{Sections/4_Method}
\input{Sections/5_Experiments}
\input{Sections/6_Discussion}

\input{Sections/7_Related_Work}
\input{Sections/8_Conclusion}

\clearpage

\bibliography{mlsys2025}
\bibliographystyle{mlsys2025}

\ifpreprint
\section*{Limitations}
\input{Sections/Limitations}
\fi

\ifpreprint
\appendix
\input{Sections/Appendix}
\fi

\end{document}

%% file: Sections/1_Introduction.tex
\begin{strip}
  \centering
  \vspace{-12pt}
  \includegraphics[width=\textwidth]{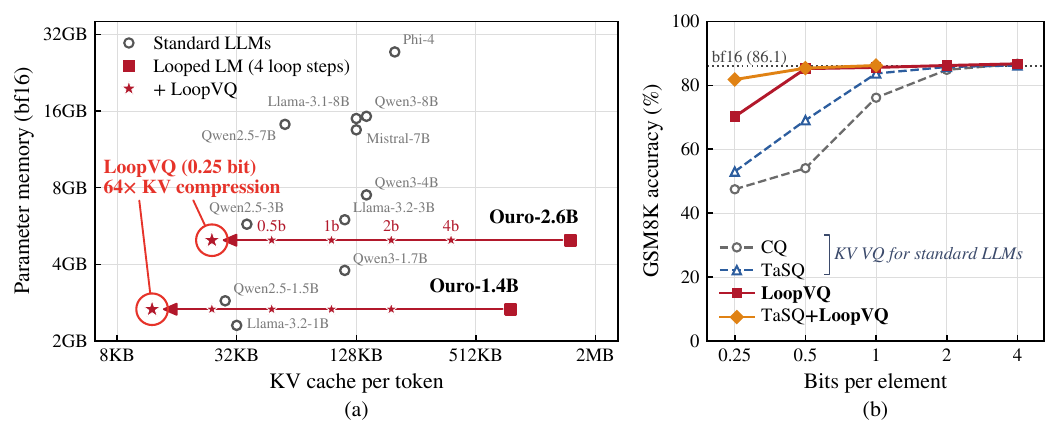}
  \vspace{-25pt}
  \captionof{figure}{\textbf{Looped LMs trade parameter memory for KV cache.}
  \textbf{(a)}~Parameter memory versus KV cache per token, both in bf16, for standard LLMs
  and the looped Ouro models with four loop steps. Arrows mark the KV cache of Ouro under LoopVQ
  from 4 to 0.25 bits per element.
  \textbf{(b)}~GSM8K accuracy of Ouro-2.6B versus the bitwidth of KV cache quantization.}
  \label{fig:motivation-overview}
\end{strip}
\section{Introduction}
\label{sec:intro}

Looped transformers~\cite{dehghani2018universal,giannou2023looped,saunshi2025reasoning,zhu2025scaling,geiping2026scaling}
apply the same layers over multiple loop steps, effectively reusing the same parameters across multiple iterations. Built on this parameter-efficient architecture, Ouro-1.4B performs on par with non-looped
models several times its size~\cite{zhu2025scaling}.

This parameter efficiency, however, does not extend to the key-value (KV) cache. Although the
loop steps share their weights, each loop step stores its own keys and values, so the KV cache
grows linearly with the number of loop steps. As Figure~\ref{fig:motivation-overview}a shows,
Ouro-1.4B requires six times the per-token KV cache of Llama-3.1-8B~\cite{2024arXiv240721783G}, and the model's KV cache exceeds its
weights in memory even at moderate context lengths. This limits the batch size and context length at which the looped models can be served, making KV cache compression a promising way to improve their efficiency.

Among various KV cache compression methods, quantization compresses the KV cache by storing each key and value at lower precision, without
discarding any tokens. It has been widely studied for standard
LLMs, with both scalar~\cite{liu2024kivi,hooper2024kvquant,ashkboos2024quarot} and vector quantization~\cite{zhang2024kv,cheong2026tailoring,son2026nsnquant}.
These methods can also be applied to looped models, but they do not exploit the redundancy of
the cache across loop steps, and below one bit per element they lose much of their accuracy
(Figure~\ref{fig:motivation-overview}b).

We find that the keys and values of a layer are strongly correlated across loop steps. Grouping
channels of several loop steps into one sub-vector exploits this correlation and substantially
reduces the reconstruction error at the same bitwidth, which makes the loop dimension an
additional axis for vector quantization, complementary to the channel axis used by existing
methods. We also find that quantization errors at different loop steps affect the model output to
very different degrees.

Based on these observations, we propose LoopVQ, a KV cache quantization method for looped
models with two components. \emph{Loop-axis vector quantization} codes channels of several loop
steps jointly, so that one index is shared by multiple loop steps with no additional cost in bits.
\emph{Loop-aware codebook fine-tuning} refines the codebooks to match the output of the
full-precision model, accounting for the unequal sensitivity of the loop steps. 

We evaluate LoopVQ on general benchmarks with Ouro-1.4B, Ouro-2.6B~\cite{zhu2025scaling}, and
Huginn-3.5B~\cite{geiping2026scaling}, and on long chain-of-thought reasoning and
needle-in-a-haystack (NIAH) tasks with the two Ouro models.
Combined with TaSQ~\cite{cheong2026tailoring}, which exploits structure along the channel axis,
LoopVQ comes close to the BF16 baseline at 0.5 bit. We further implement LoopVQ in
SGLang~\cite{zheng2024sglang}, where the loop steps share the same indices and each reads them
through its own slice of the codebook.

Our contributions are as follows:
\begin{itemize}
  \item We show that the KV cache of looped models is highly redundant along the loop axis and
  that the loop steps differ substantially in their sensitivity to quantization.
  \item We propose LoopVQ, which combines loop-axis vector quantization with loop-aware codebook
  fine-tuning and is orthogonal to channel-axis methods such as TaSQ. At 0.5 and 0.25 bits per
  element, LoopVQ outperforms vector quantization baselines designed for non-looped LLMs on
  general, reasoning, and long-context benchmarks.
  \item We implement LoopVQ in SGLang, where the quantized cache stores one index row for several loop steps. It
  nearly halves the decode latency at long contexts and increases the peak throughput by up to
  $4.3\times$ and $6.1\times$ on the 1.4B and 2.6B Ouro models compared to the BF16 baseline.
\end{itemize}

%% file: Sections/2_Preliminaries.tex
\section{Preliminaries}
\label{sec:prelim}

\paragraph{Looped language models.}
A looped transformer~\cite{zhu2025scaling,geiping2026scaling} applies the same $L_b$ decoder
layers $T$ times, where the first layer of each loop step takes the previous loop step's output as its input. Starting from the embedded sequence $h^{(0)}_{1:N}$, loop step $t\in\{1,\dots,T\}$
computes
\begin{equation}
  h^{(t)}_{1:N} \;=\; F_\theta\bigl(h^{(t-1)}_{1:N}\bigr),
  \label{eq:loop}
\end{equation}
where $F_\theta$ denotes the $L_b$ layers, shared across loop steps. The output distribution $p_j$ is
read from the final hidden state $h^{(T)}_j$. Each layer stores a separate KV
cache at every loop step. Token $j$ produces a key $k^{(t,\ell)}_{j,h}\in\mathbb{R}^D$ and a value
$v^{(t,\ell)}_{j,h}\in\mathbb{R}^D$ for each loop step $t$, layer $\ell$ and KV head $h\le H$.
The cache is $T$ times larger than that of a
non-looped model with the same weights.

\paragraph{Vector quantization of the KV cache.}
Prior work compresses the KV cache with vector quantization~\cite{zhang2024kv}. Every cached vector
$x\in\mathbb{R}^D$ ($x=k$ or $v$ of one loop step, layer, and head) is split into $D/G$ sub-vectors
$z\in\mathbb{R}^{G}$ of $G$ contiguous channels. Each sub-vector is stored as the index of its
nearest entry in a codebook $\mathcal{B}=\{e_1,\dots,e_K\}\subset\mathbb{R}^G$ of $K=2^{b}$
entries,
\begin{equation}
  a(z)\;=\;\arg\min_{r\le K}\;\|z-e_r\|_2^2,\qquad \hat z=e_{a(z)} ,
  \label{eq:nn}
\end{equation}
at a cost of $b/G$ bits per element. Keys are quantized before the rotary embedding (RoPE), so
attention reconstructs each key from its indices and applies RoPE to it, reading $D/G$ indices per
cached vector. When these methods are applied to a looped model, codebooks are learned per loop step and layer $(t,\ell)$, and the
$T$ loop steps of a layer are quantized as unrelated caches.

\begin{figure*}[t!]
  \centering
  \includegraphics[width=0.9\textwidth]{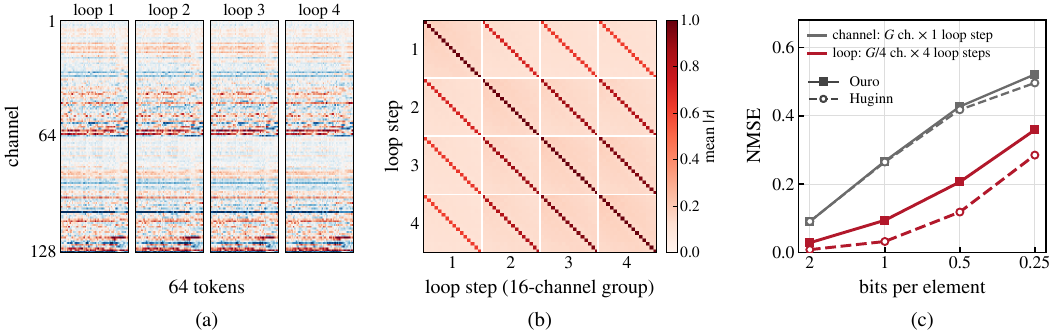}
  \caption{\textbf{The KV cache of a looped LM is redundant along the loop axis.}
  \textbf{(a)}~Pre-RoPE keys of one head of Ouro-1.4B at each loop step.
  \textbf{(b)}~Mean absolute correlation between (loop step, channel) pairs within a
  16-channel group, averaged over all groups, heads, and layers of Ouro-1.4B.
  \textbf{(c)}~$k$-means reconstruction error at equal bitwidth when each 8-bit code
  spans $G$ channels of one loop step (channel) or $G/4$ channels across four loop steps (loop),
  both with plain $k$-means.}
  \label{fig:loop_redundancy}
\end{figure*}

\paragraph{Codebook calibration.}
Codebooks are learned offline from sub-vectors $\{z_n\}$ collected on a small calibration set,
typically by $k$-means on their squared reconstruction error. This objective treats all
sub-vectors alike, although the error in some may affect the model output more than others. To address this, prior work weights each sub-vector by its sensitivity
$\omega_n=\|\partial\mathcal{L}/\partial z_n\|_2^2$ to the next-token loss $\mathcal{L}$, which
equals the sum of the diagonal empirical Fisher information over its coordinates:
\begin{equation}
  \min_{\mathcal{B}}\;\sum_n \omega_n\,\min_{r\le K}\bigl\|z_n-e_r\bigr\|_2^2 .
  \label{eq:wkmeans}
\end{equation}
This gradient-based weighting has been reported to improve over unweighted $k$-means in KV cache
quantization~\cite{zhang2024kv,hooper2024kvquant}. It only shapes the codebooks; encoding remains the nearest-entry
rule~\eqref{eq:nn}.

%% file: Sections/3_Motivation.tex
\section{Motivation}
\label{sec:motivation}

Unlike the caches of different layers in a non-looped model, the $T$ caches of a layer in a
looped model are produced by applying the same parameters. We investigate whether this property can be used
to compress the cache beyond what is achievable for a non-looped model.

\paragraph{Redundancy along the loop axis.}
The keys and values of a layer are highly similar across loop steps, as they are computed by the
same projection weights from hidden states of the same token~\eqref{eq:loop}. In
Figure~\ref{fig:loop_redundancy}a, the keys of one head show nearly the same channel pattern at
all four loop steps. Averaged over all layers and heads of Ouro-1.4B, the correlation between
adjacent loop steps increases from $0.69$ between the first two loop steps to $0.94$ between the
last two.

This redundancy runs along a different axis from the one that existing vector quantization methods exploit.
Figure~\ref{fig:loop_redundancy}b shows the correlation between every pair of (loop step,
channel) coordinates within a 16-channel sub-vector. A key channel correlates with the same
channel at the adjacent loop step at $|r|=0.85$ on average, but with the other channels of its
own sub-vector at only $0.16$, and the gap is even wider for values.
Figure~\ref{fig:loop_redundancy}c measures how much this matters for quantization. At the same bitwidth and with the same
number of codebooks, we only change which coordinates share a code: $G$ channels of one loop step, or
$G/4$ channels at each of four consecutive loop steps. With plain $k$-means, the second grouping
roughly halves the reconstruction error of Ouro-1.4B at 0.5 bit, and reduces it substantially
for both models at every bitwidth.

\begin{table}[t]
  \centering
  \small
  \setlength{\tabcolsep}{5pt}
  \begin{tabular}{lcccc}
    \toprule
    & \multicolumn{4}{c}{Quantized loop step} \\
    \cmidrule(lr){2-5}
    Bitwidth & 1 & 2 & 3 & 4 \\
    \midrule
    0.5 bit  & 0.069 & 0.018 & 0.013 & 0.030 \\
    0.25 bit & 0.123 & 0.029 & 0.026 & 0.065 \\
    \bottomrule
  \end{tabular}
  \caption{KL divergence from the full-precision output of Ouro-1.4B when only one loop step is
  quantized (loop-axis grouping, Fisher-weighted $k$-means codebooks, sequences not used for calibration).}
  \label{tab:step_kl}
\end{table}

\begin{figure*}[t!]
  \centering
  \includegraphics[width=\textwidth]{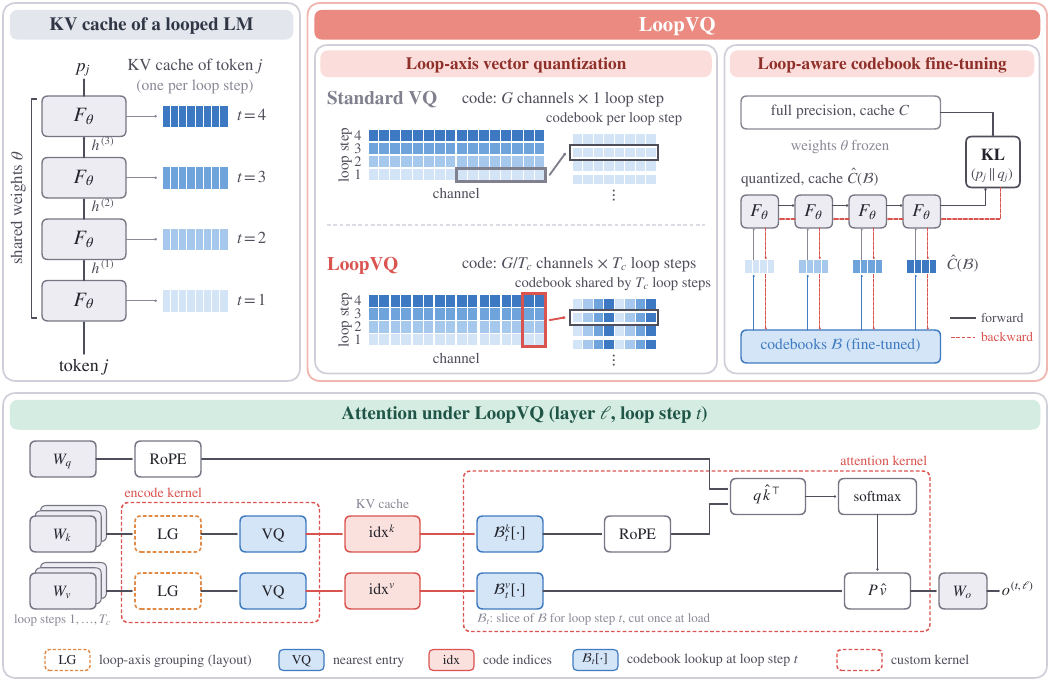}
  \caption{\textbf{Overview of LoopVQ.} Each loop step of a looped LM writes its own KV cache.
  LoopVQ codes $G/T_c$ channels across $T_c$ loop steps, so one index serves all of them at the
  same bitwidth as coding each loop step separately, and fine-tunes the codebooks against the full-precision output
  through the loop. At inference, each loop step reads the shared indices through its own slice
  $\mathcal{B}_t$ of the codebook.}
  \label{fig:method_overview}
\end{figure*}

\paragraph{Unequal quantization sensitivity across loop steps.}
Grouping along the loop axis places coordinates from different loop steps in one sub-vector,
which is then encoded by a single index. However, the loop steps do not affect the output
equally. An error in the cache of an early loop step perturbs the hidden state passed to all later loop
steps~\eqref{eq:loop}, whereas an error at the last loop step affects only that loop step.

Table~\ref{tab:step_kl} confirms this. We quantize the cache of a single loop step with
loop-axis codebooks calibrated by~\eqref{eq:wkmeans}, keep all other loop steps in full precision,
and measure the KL divergence of the output distribution from that of the full-precision model.
The sensitivity varies by roughly a factor of five across loop steps: quantizing the first loop step is by
far the most harmful, and the third loop step the least.

The calibration objective~\eqref{eq:wkmeans} cannot express this difference within a coupled
sub-vector. Its weight $\omega_n$ is a single scalar per sub-vector, and its distance is the
unweighted squared error, so an error is penalized equally regardless of the loop step on which
it falls. The codebooks allocate the same precision to every loop step of a sub-vector,
although errors at some loop steps affect the output far more than errors at others.

Existing KV cache quantization methods do not account for two properties of the looped structure: the cache is highly redundant along the loop axis, and the loop steps differ
substantially in how strongly their quantization errors affect the output.
Section~\ref{sec:method} builds on both observations.

%% file: Sections/4_Method.tex
\section{Method}
\label{sec:method}

LoopVQ builds on the two properties identified in Section~\ref{sec:motivation}: the redundancy of
the cache along the loop axis (Section~\ref{sec:loopvq}) and the unequal sensitivity of the loop
steps (Section~\ref{sec:ft}). Figure~\ref{fig:method_overview} gives an overview.

\subsection{Loop-axis vector quantization}
\label{sec:loopvq}

LoopVQ partitions the $T$ loop steps into blocks of $T_c$ consecutive loop steps, where $T_c$ divides
both $T$ and $G$; block $\beta$ contains the loop steps $\beta T_c+m+1$ with $m<T_c$. A code then covers
$c=G/T_c$ channels at each of the $T_c$ loop steps of a block. For $x\in\{k,v\}$, layer $\ell$, head $h$,
and channel group $g<D/c$, the coupled vector of token $j$ is
\begin{equation}
  \bigl[z^{(\beta,\ell,h,g)}_j\bigr]_{iT_c+m}
  = x^{(\beta T_c+m+1,\,\ell)}_{j,h,\,gc+i},
  \quad i<c,\; m<T_c .
  \label{eq:coupled}
\end{equation}
As in existing KV cache vector quantization methods, LoopVQ quantizes the keys before RoPE is applied, and attention applies RoPE to the reconstructed keys (Section~\ref{sec:prelim}). The
coupled vector is encoded by its nearest entry~\eqref{eq:nn} in a codebook
$\mathcal{B}_{\beta,\ell,h,g}\subset\mathbb{R}^{G}$, and loop step $\beta T_c+m+1$ is reconstructed from
its share of that entry:
\begin{equation}
  \hat x^{(\beta T_c+m+1,\,\ell)}_{j,h,\,gc+i}\;=\;\bigl[e_{a(z_j)}\bigr]_{iT_c+m}.
  \label{eq:decode}
\end{equation}
One index serves $T_c$ loop steps, and the bitwidth remains $b/G$ bits per element. The codebooks
are initialized with Fisher-weighted $k$-means~\eqref{eq:wkmeans}, where
$\omega_n=\|\partial\mathcal{L}/\partial z_n\|_2^2$ sums the squared gradients over all $T_c$ loop steps
covered by $z_n$.

\begin{figure}[t]
  \centering
  \includegraphics[width=0.85\columnwidth]{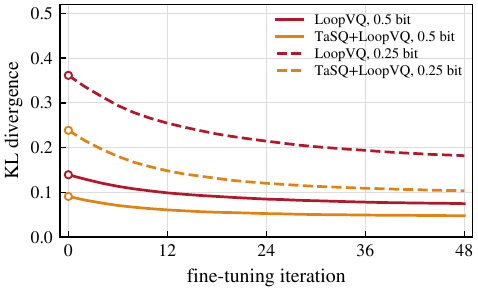}
  \caption{KL divergence from the full-precision Ouro-1.4B during fine-tuning, starting from the
  $k$-means codebooks (circles). One iteration processes one calibration sequence (48 iterations = 4 epochs).}
  \label{fig:ft_kl}
\end{figure}

\subsection{Loop-aware codebook fine-tuning}
\label{sec:ft}

We refine the $k$-means codebooks by matching the output of the quantized model to that of the
original model, as in knowledge distillation~\cite{hinton2015distilling} and the end-to-end
fine-tuning of vector-quantized LLM weights~\cite{egiazarian2024extreme,malinovskii2024pv}. For a calibration sequence $s$, let $C$ be the cache of the unquantized forward
pass and $p_j$ its output distribution at position $j$. Let $\hat C(\mathcal{B})$ be the cache
reconstructed from the codes of $C$ by~\eqref{eq:nn} and~\eqref{eq:decode}, where $\mathcal{B}$
collects all key and value codebooks, and let $q_j(\mathcal{B})$ be the output when all $T$ loop
steps attend to $\hat C(\mathcal{B})$ instead of $C$. With the model weights frozen, we minimize
\begin{equation}
  \mathcal{J}(\mathcal{B}) \;=\; \mathbb{E}_{s}\Bigl[\tfrac{1}{N}\textstyle\sum_{j}
  \mathrm{KL}\bigl(p_j\,\big\|\,q_j(\mathcal{B})\bigr)\Bigr],
  \label{eq:ft}
\end{equation}
where the sum runs over the $N$ positions after the bf16 sink tokens.

Fisher-weighted $k$-means~\eqref{eq:wkmeans}, like any weighted $k$-means objective, penalizes the
error of each coded vector independently. It cannot account for the interaction between
the errors of different loop steps, which propagate to later loop steps and combine
(Appendix~\ref{app:ft-fisher}); its weights also treat all $T_c$ loop steps of a coupled vector
equally, although they differ in sensitivity (Section~\ref{sec:motivation}). Minimizing~\eqref{eq:ft}
retains both effects. On Ouro-1.4B, fine-tuning
roughly halves the KL divergence from the full-precision output (Figure~\ref{fig:ft_kl}).

The entries are trained by backpropagating~\eqref{eq:ft} through the reconstruction
(Appendix~\ref{app:ft-grad}). We recompute the assignments with the current codebooks at every
iteration, so the codebooks are always trained under the encoder used at inference.

\begin{figure*}[t]
  \centering
  \includegraphics[width=\textwidth]{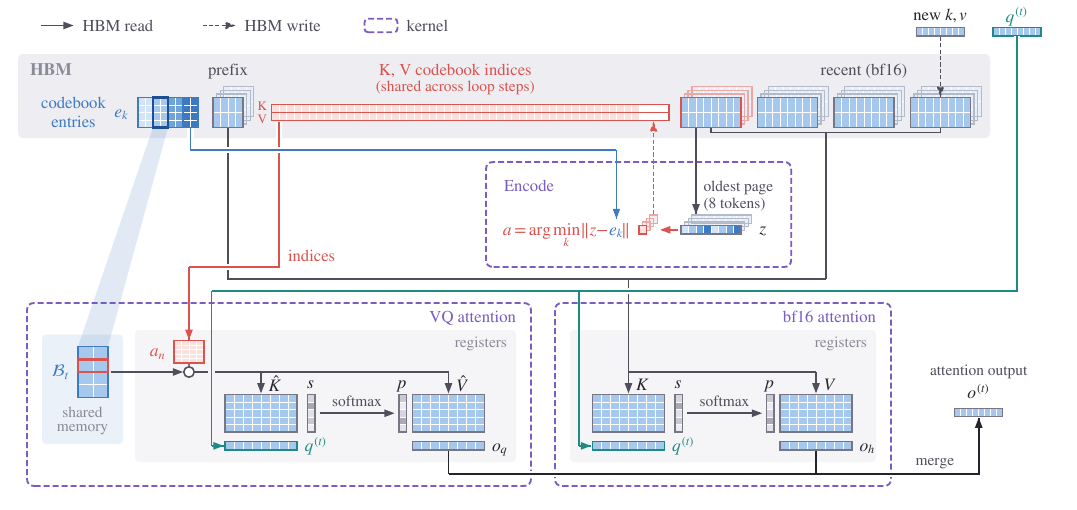}
  \caption{\textbf{Serving LoopVQ in SGLang.} The KV cache keeps the sink and recent tokens in
  bf16 at every loop step, and all other tokens as codebook indices shared across loop steps. At loop step $t$, the LoopVQ attention kernel loads $\mathcal{B}_t$, the slice of the
  codebook entries for loop step $t$, into shared memory once and looks up the keys and values of
  each token from its indices in registers; the bf16 kernel reads the bf16 keys and values of
  loop step $t$ directly. Both attend with the query $q^{(t)}$ of the new token, and their outputs are merged.
  The key and value of each new token are written in bf16, and after every 8 generated tokens the
  oldest recent page is encoded into indices.}
  \label{fig:serving}
\end{figure*}

\begin{algorithm}[t]
  \caption{\textbf{LoopVQ attention kernel.} Decode attention at loop step $t$ for one head and one
  split $\mathcal{S}$.}
  \label{alg:vq_attention}
  \begin{algorithmic}[1]
    \REQUIRE query $q^{(t)}$; codebook slices $\mathcal{B}^k_t$, $\mathcal{B}^v_t$; key and value
    indices $a^k_n$, $a^v_n$ of each token $n\in\mathcal{S}$
    \ENSURE partial output $o$ and its logsumexp $\lambda$
    \STATE load $\mathcal{B}^k_t$ into shared memory
    \FOR{$n\in\mathcal{S}$ \textbf{in parallel}}
      \STATE $\hat k_n\gets\mathcal{B}^k_t[a^k_n]$ \COMMENT{$D/c$ indices}
      \STATE $s_n\gets q^{(t)}\cdot\mathrm{RoPE}(\hat k_n,n)/\sqrt{D}$
    \ENDFOR
    \STATE $\mu\gets\max_n s_n$, \; $p_n\gets e^{s_n-\mu}$, \; $Z\gets\sum_n p_n$
    \STATE load $\mathcal{B}^v_t$ into shared memory
    \STATE $o\gets\frac{1}{Z}\sum_{n\in\mathcal{S}}p_n\,\mathcal{B}^v_t[a^v_n]$
    \STATE \textbf{return} $o$, \; $\lambda=\mu+\log Z$
  \end{algorithmic}
\end{algorithm}

\subsection{System implementation}
\label{sec:serving}

We implement LoopVQ in SGLang~\cite{zheng2024sglang} (Figure~\ref{fig:serving}). Coupling loop
steps affects how the codes are stored, when a token can be encoded, and how attention at one loop
step reads codes shared by several loop steps.

\paragraph{Cache layout.}
The first $n_s$ tokens (sink) and the most recent $n_r$ tokens of a request are kept in bf16,
and all other tokens are stored as codebook indices. The cache is organized in pages of 8
tokens, and $n_s$ and $n_r$ are multiples of 8, so every page is either bf16 or quantized as a
whole. Since one index is shared across $T_c$ loop steps~\eqref{eq:decode}, a quantized page stores one
index row per block of $T_c$ loop steps, and LoopVQ occupies the same memory
per token as coding each loop step separately at the same $G$.

\paragraph{Encoding.}
A coupled vector~\eqref{eq:coupled} can be encoded only once all $T_c$ loop steps of its block have been
computed. During prefill, the keys and values of each loop step are held until
the last loop step of their block, and the block is then encoded by computing the distances to all
codebook entries as a matrix product on tensor cores. During
decoding, the key and value of each new token are written to the bf16 recent window, where all
its loop steps are available by the end of the forward pass. After every 8 generated tokens, the
oldest page of the window is encoded across all layers and loop steps into one quantized page, so
the window holds $n_r$ to $n_r+7$ tokens.

\paragraph{Attention.}
At loop step $t$, one kernel attends to the bf16 pages and another to the quantized pages
(Algorithm~\ref{alg:vq_attention}). The quantized kernel looks up each index in $\mathcal{B}_t$, the slice of every entry that holds the
coordinates of loop step $t=\beta T_c+m+1$, i.e., $\bigl[e\bigr]_{iT_c+m}$ for $i<c$~\eqref{eq:decode},
cut once when the model is loaded. Each loop step uses an ordinary vector quantizer with groups
of $c$ channels, and the baselines, which code each loop step separately, run on the same kernel. Keys are cached before the rotary
embedding, which the kernel applies to the reconstructed keys. At each loop step, the kernel
reads $D/c$ indices per head and token. As in Flash-Decoding~\cite{dao2023flashdecoding}, the
quantized tokens are divided into splits processed by separate thread blocks, and the partial
outputs of all splits are combined using the logsumexp of their attention scores~\cite{dao2024flashattention}, which recovers
the exact softmax over all tokens. The bf16 pages are attended by SGLang's Triton decoding kernel,
which uses the same scheme, and its splits are combined with those of the quantized kernel. For
batched decoding, the quantized kernel is optimized to process multiple requests per thread block, sharing
the loads of the rotary coefficients and codebooks (Appendix~\ref{app:kernel}).

%% file: Sections/5_Experiments.tex
\begin{table*}[t!]
\centering
\caption{General performance under low-bit KV cache quantization. All results use greedy decoding. Bold denotes the best quantized result; higher is better.}
\label{tab:general_results}
\resizebox{\linewidth}{!}{\input{Tables/general_results}}
\end{table*}

\begin{table*}[t!]
\centering
\caption{Long CoT reasoning performance under low-bit KV cache quantization. All results are averaged over three sampling seeds (mean $\pm$ std). Bold denotes the best quantized result; higher is better.}
\label{tab:reasoning_results}
\resizebox{\linewidth}{!}{\input{Tables/reasoning_results}}
\end{table*}

\section{Experiments}
\label{sec:experiments}

\subsection{Experimental setup}
\label{sec:setup}

\paragraph{Models.}
We evaluate Ouro-1.4B and Ouro-2.6B~\cite{zhu2025scaling} ($L_b=24$ and $48$, $T=4$, $H=16$,
$D=128$), using their Thinking variants for reasoning tasks, and Huginn-3.5B~\cite{geiping2026scaling},
which iterates a four-layer core block $r=32$ times between a two-layer prelude and coda
(132 KV caches per token, 55 heads of dimension 96).

\paragraph{Methods.}
All methods quantize keys and values with 8-bit codes ($K=256$) and $G\in\{16,32\}$, i.e., 0.5 or
0.25 bits per element. The baselines are CQ~\cite{zhang2024kv}, with a separate Fisher-weighted $k$-means
codebook~\eqref{eq:wkmeans} per loop step and layer, and TaSQ~\cite{cheong2026tailoring}, which transforms the
keys before coding each loop step. LoopVQ couples $T_c=4$ consecutive loop steps, so a code spans $c=G/T_c$
channels ($c=4$ at 0.5 bit, $c=8$ at 0.25 bit); on Huginn, the 32 loop steps form eight blocks,
and the prelude and coda, which have no loop axis, are coded per layer as in CQ. TaSQ+LoopVQ
applies TaSQ's key transforms before coupling (Appendix~\ref{app:tasq}); its per-token key scale
adds 0.008 bits per key element (0.003 on Huginn). Every method keeps the first and the most recent tokens in bf16:
8 and 64 on the general benchmarks, 64 and 256 on the reasoning and long-context benchmarks.

\paragraph{Calibration.}
All codebooks are fitted separately for each model, including the Thinking variants, on 16 sequences of 2,048 tokens from the GPQA main set excluding the Diamond subset,
which does not overlap with the evaluation set. LoopVQ fine-tunes the $k$-means codebooks on 12 of
these sequences with Adam (learning rate $10^{-3}$, one sequence per iteration) for four epochs and
keeps the epoch with the lowest KL on the other four. On one RTX~6000 Ada GPU, the $k$-means calibration and the fine-tuning take 6.1 and 2.3
minutes for Ouro-1.4B and 14.3 and 5.9 minutes for Ouro-2.6B.

\paragraph{Benchmarks.}
We evaluate three settings, all with accuracy (\%) as the metric.
\emph{General benchmarks}: the base models are evaluated with greedy decoding on
GSM8K~\cite{cobbe2021training} (3-shot chain of thought, flexible extraction),
MATH500~\cite{lightman2024let} (zero-shot), and HumanEval~\cite{chen2021evaluating} and
MBPP~\cite{austin2021program} (pass@1).
\emph{Long chain-of-thought reasoning}: the Thinking models are evaluated with thinking enabled on
AIME'24, AIME'25~\cite{aime}, GPQA-Diamond~\cite{rein2023gpqa},
HumanEval+, and MBPP+~\cite{liu2023your}, with the decoding settings of the model card (temperature 1.0,
top-$p$ 0.7, up to 32,768 generated tokens), averaged over three sampling seeds.
\emph{Long-context retrieval}: the base models are evaluated on the eight needle-in-a-haystack (NIAH) tasks
of RULER~\cite{hsieh2024ruler} at context lengths from 4K to 64K tokens, the maximum context length
of Ouro, with 200 examples per length, averaged over three seeds for the haystack and needle
placement.

\begin{figure*}[t]
  \centering
  \includegraphics[width=\textwidth]{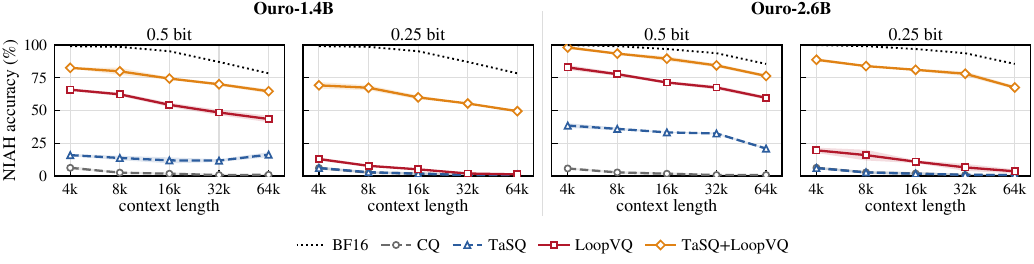}
  \caption{RULER needle-in-a-haystack (NIAH) accuracy of Ouro-1.4B and Ouro-2.6B versus context length under 0.5- and 0.25-bit KV cache quantization (8-bit codes, $G{=}16$ and $32$).
  Each point averages the eight NIAH tasks over three seeds.}
  \label{fig:niah}
\end{figure*}

\paragraph{Implementation and hardware.}
All methods run in SGLang~\cite{zheng2024sglang} and store the quantized tokens as code
indices. Decoding attends to the codes through a custom attention kernel; prefill attends
to the exact keys and values of the current chunk and encodes each block of $T_c$ loop steps once its
last loop step is computed.
Efficiency is measured on an NVIDIA RTX~6000 Ada GPU (48\,GB), and results on A100 (40\,GB) are
reported in Appendix~\ref{app:efficiency}.

\subsection{Main results}

\paragraph{General benchmarks.}
Table~\ref{tab:general_results} reports the results on the base models. CQ and TaSQ lose much of
the BF16 accuracy already at 0.5 bit, especially on MATH500 and HumanEval. LoopVQ substantially
outperforms both baselines at both bitwidths, and TaSQ+LoopVQ further improves on LoopVQ, nearly
matching BF16 at 0.5 bit. At 0.25 bit, the advantage of TaSQ+LoopVQ over LoopVQ becomes larger.
LoopVQ and TaSQ+LoopVQ also outperform both baselines on Huginn-3.5B.

\paragraph{Long chain-of-thought reasoning.}
Table~\ref{tab:reasoning_results} reports the results on the Thinking models. The baselines degrade
more severely than on the general benchmarks and fall far below BF16 at both bitwidths. LoopVQ retains most of the
BF16 accuracy at 0.5 bit, and combined with TaSQ it preserves much of the reasoning ability even at
0.25 bit, achieving the best result on every benchmark.

\paragraph{Long-context retrieval.}
Figure~\ref{fig:niah} shows the NIAH results, which follow the same
ordering. The baselines remain far below BF16 at every context length, while LoopVQ retrieves a large fraction of
the needles at 0.5 bit. TaSQ+LoopVQ achieves the highest accuracy at every context length and bitwidth.
Results on LongBench, which follow the same trend, are reported in Appendix~\ref{app:longbench}.

\begin{figure*}[t]
  \centering
  \includegraphics[width=\textwidth]{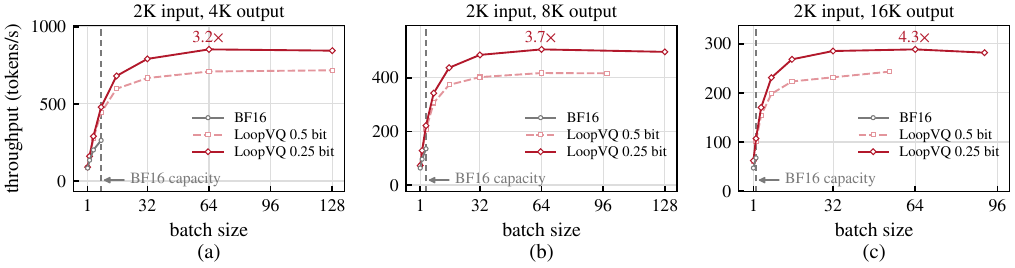}
  \vspace{-20pt}
  \caption{\textbf{End-to-end throughput of LoopVQ} on Ouro-1.4B-Thinking with one RTX~6000 Ada
  GPU, versus batch size up to the largest batch that fits in memory.}
  \label{fig:throughput}
\end{figure*}

\begin{figure}[t]
  \centering
  \includegraphics[width=\columnwidth]{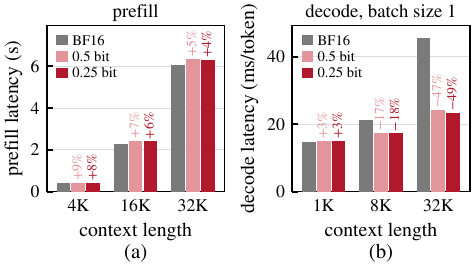}
  \vspace{-20pt}
  \caption{\textbf{Latency of LoopVQ} at 0.5 and 0.25 bit on Ouro-1.4B-Thinking with one RTX~6000
  Ada GPU. (a)~Prefill latency and (b)~decode latency per token at batch size 1, each with the change
  relative to BF16.}
  \label{fig:latency}
\end{figure}

\subsection{System efficiency}
\label{sec:efficiency}

We measure the serving efficiency of LoopVQ on Ouro-1.4B-Thinking with one RTX~6000 Ada GPU,
keeping 8 sink and 64 recent tokens in bf16; results on an A100 GPU are reported in
Appendix~\ref{app:efficiency}.

\paragraph{Prefill and decoding latency.}
Prefill with LoopVQ is slightly slower than with BF16, because of the encoding of each block
of $T_c$ loop steps, and the overhead decreases with the context length
(Figure~\ref{fig:latency}a). During decoding, the kernel reads $D/c$ indices per head and token at
each loop step, $8\times$ and $16\times$ fewer bytes than the bf16 keys and values. As decoding at
long contexts is bound by reading the cache, the reduction in decode latency grows with the context
length, reaching about half of the BF16 latency at 32K tokens (Figure~\ref{fig:latency}b). Appendix~\ref{app:profile} breaks down the GPU time by kernel
type.

\paragraph{Batched throughput.}
For long generation, the batch size is limited by the KV cache (Figure~\ref{fig:throughput}). With BF16, only a few requests fit
in memory, while LoopVQ can serve tens to over a hundred concurrent requests. At the same batch size LoopVQ is also faster, for the same reason as its lower decode latency. As a result, the peak throughput of LoopVQ exceeds that of BF16 by $3.2\times$ to $4.3\times$, with the gap widening as the output grows longer and the BF16 batch shrinks. On Ouro-2.6B-Thinking,
the BF16 cache holds even fewer requests, and the gain reaches $5.0\times$ to $6.1\times$
(Appendix~\ref{app:efficiency_2p6b}).

%% file: Tables/general_results.tex
\setlength{\aboverulesep}{0pt}\setlength{\belowrulesep}{0pt}\renewcommand{\arraystretch}{1.15}
\begin{tabular}{l|llcrrrrr}
\toprule
Model & Bitwidth & Method & Bits (K/V) & GSM8K & MATH500 & HumanEval & MBPP & Avg. \\
\midrule
\multirow{9}{*}{Ouro-1.4B} &  & BF16 & 16/16 & 78.77 & 77.20 & 73.17 & 73.54 & 75.67 \\
\cmidrule(r){2-9}
 & \multirow{4}{*}{0.5 bit} & CQ & 0.5/0.5 & 46.78 & 17.80 & 14.63 & 50.00 & 32.30 \\
 &  & TaSQ & 0.508/0.5 & 57.32 & 38.80 & 24.39 & 53.97 & 43.62 \\
 &  & \cellcolor[HTML]{E8F1FB}\textbf{LoopVQ} & \cellcolor[HTML]{E8F1FB}0.5/0.5 & \cellcolor[HTML]{E8F1FB}75.82 & \cellcolor[HTML]{E8F1FB}69.40 & \cellcolor[HTML]{E8F1FB}63.41 & \cellcolor[HTML]{E8F1FB}64.55 & \cellcolor[HTML]{E8F1FB}68.29 \\
 &  & \cellcolor[HTML]{E8F1FB}TaSQ+\textbf{LoopVQ} & \cellcolor[HTML]{E8F1FB}0.508/0.5 & \cellcolor[HTML]{E8F1FB}\textbf{77.56} & \cellcolor[HTML]{E8F1FB}\textbf{75.20} & \cellcolor[HTML]{E8F1FB}\textbf{70.12} & \cellcolor[HTML]{E8F1FB}\textbf{71.43} & \cellcolor[HTML]{E8F1FB}\textbf{73.58} \\
\cmidrule(r){2-9}
 & \multirow{4}{*}{0.25 bit} & CQ & 0.25/0.25 & 42.99 & 12.20 & 10.98 & 46.56 & 28.18 \\
 &  & TaSQ & 0.258/0.25 & 44.88 & 12.80 & 12.20 & 48.41 & 29.57 \\
 &  & \cellcolor[HTML]{E8F1FB}\textbf{LoopVQ} & \cellcolor[HTML]{E8F1FB}0.25/0.25 & \cellcolor[HTML]{E8F1FB}58.91 & \cellcolor[HTML]{E8F1FB}41.20 & \cellcolor[HTML]{E8F1FB}25.61 & \cellcolor[HTML]{E8F1FB}54.50 & \cellcolor[HTML]{E8F1FB}45.05 \\
 &  & \cellcolor[HTML]{E8F1FB}TaSQ+\textbf{LoopVQ} & \cellcolor[HTML]{E8F1FB}0.258/0.25 & \cellcolor[HTML]{E8F1FB}\textbf{70.96} & \cellcolor[HTML]{E8F1FB}\textbf{63.40} & \cellcolor[HTML]{E8F1FB}\textbf{55.49} & \cellcolor[HTML]{E8F1FB}\textbf{59.26} & \cellcolor[HTML]{E8F1FB}\textbf{62.28} \\
\midrule
\multirow{9}{*}{Ouro-2.6B} &  & BF16 & 16/16 & 86.13 & 77.40 & 81.10 & 81.22 & 81.46 \\
\cmidrule(r){2-9}
 & \multirow{4}{*}{0.5 bit} & CQ & 0.5/0.5 & 54.06 & 24.60 & 14.63 & 57.41 & 37.67 \\
 &  & TaSQ & 0.508/0.5 & 69.07 & 49.20 & 31.10 & 61.38 & 52.69 \\
 &  & \cellcolor[HTML]{E8F1FB}\textbf{LoopVQ} & \cellcolor[HTML]{E8F1FB}0.5/0.5 & \cellcolor[HTML]{E8F1FB}85.22 & \cellcolor[HTML]{E8F1FB}74.60 & \cellcolor[HTML]{E8F1FB}75.00 & \cellcolor[HTML]{E8F1FB}75.93 & \cellcolor[HTML]{E8F1FB}77.69 \\
 &  & \cellcolor[HTML]{E8F1FB}TaSQ+\textbf{LoopVQ} & \cellcolor[HTML]{E8F1FB}0.508/0.5 & \cellcolor[HTML]{E8F1FB}\textbf{85.37} & \cellcolor[HTML]{E8F1FB}\textbf{78.60} & \cellcolor[HTML]{E8F1FB}\textbf{78.66} & \cellcolor[HTML]{E8F1FB}\textbf{78.57} & \cellcolor[HTML]{E8F1FB}\textbf{80.30} \\
\cmidrule(r){2-9}
 & \multirow{4}{*}{0.25 bit} & CQ & 0.25/0.25 & 47.46 & 14.60 & 12.80 & 53.70 & 32.14 \\
 &  & TaSQ & 0.258/0.25 & 52.99 & 17.60 & 14.02 & 56.08 & 35.18 \\
 &  & \cellcolor[HTML]{E8F1FB}\textbf{LoopVQ} & \cellcolor[HTML]{E8F1FB}0.25/0.25 & \cellcolor[HTML]{E8F1FB}70.20 & \cellcolor[HTML]{E8F1FB}52.00 & \cellcolor[HTML]{E8F1FB}28.66 & \cellcolor[HTML]{E8F1FB}60.05 & \cellcolor[HTML]{E8F1FB}52.73 \\
 &  & \cellcolor[HTML]{E8F1FB}TaSQ+\textbf{LoopVQ} & \cellcolor[HTML]{E8F1FB}0.258/0.25 & \cellcolor[HTML]{E8F1FB}\textbf{81.80} & \cellcolor[HTML]{E8F1FB}\textbf{68.60} & \cellcolor[HTML]{E8F1FB}\textbf{62.80} & \cellcolor[HTML]{E8F1FB}\textbf{69.58} & \cellcolor[HTML]{E8F1FB}\textbf{70.70} \\
\midrule
\multirow{9}{*}{Huginn-3.5B} &  & BF16 & 16/16 & 34.04 & 12.80 & 17.07 & 35.98 & 24.97 \\
\cmidrule(r){2-9}
 & \multirow{4}{*}{0.5 bit} & CQ & 0.5/0.5 & 19.03 & 5.40 & 6.71 & 23.54 & 13.67 \\
 &  & TaSQ & 0.503/0.5 & 19.11 & 5.80 & 10.37 & 23.81 & 14.77 \\
 &  & \cellcolor[HTML]{E8F1FB}\textbf{LoopVQ} & \cellcolor[HTML]{E8F1FB}0.5/0.5 & \cellcolor[HTML]{E8F1FB}32.37 & \cellcolor[HTML]{E8F1FB}9.40 & \cellcolor[HTML]{E8F1FB}\textbf{12.80} & \cellcolor[HTML]{E8F1FB}34.92 & \cellcolor[HTML]{E8F1FB}22.37 \\
 &  & \cellcolor[HTML]{E8F1FB}TaSQ+\textbf{LoopVQ} & \cellcolor[HTML]{E8F1FB}0.503/0.5 & \cellcolor[HTML]{E8F1FB}\textbf{33.13} & \cellcolor[HTML]{E8F1FB}\textbf{12.20} & \cellcolor[HTML]{E8F1FB}12.20 & \cellcolor[HTML]{E8F1FB}\textbf{36.51} & \cellcolor[HTML]{E8F1FB}\textbf{23.51} \\
\cmidrule(r){2-9}
 & \multirow{4}{*}{0.25 bit} & CQ & 0.25/0.25 & 19.11 & 7.20 & 4.27 & 21.16 & 12.93 \\
 &  & TaSQ & 0.253/0.25 & 19.18 & 6.60 & 4.88 & 21.16 & 12.96 \\
 &  & \cellcolor[HTML]{E8F1FB}\textbf{LoopVQ} & \cellcolor[HTML]{E8F1FB}0.25/0.25 & \cellcolor[HTML]{E8F1FB}25.02 & \cellcolor[HTML]{E8F1FB}8.20 & \cellcolor[HTML]{E8F1FB}\textbf{13.41} & \cellcolor[HTML]{E8F1FB}27.78 & \cellcolor[HTML]{E8F1FB}18.60 \\
 &  & \cellcolor[HTML]{E8F1FB}TaSQ+\textbf{LoopVQ} & \cellcolor[HTML]{E8F1FB}0.253/0.25 & \cellcolor[HTML]{E8F1FB}\textbf{28.13} & \cellcolor[HTML]{E8F1FB}\textbf{9.80} & \cellcolor[HTML]{E8F1FB}\textbf{13.41} & \cellcolor[HTML]{E8F1FB}\textbf{32.54} & \cellcolor[HTML]{E8F1FB}\textbf{20.97} \\
\bottomrule
\end{tabular}

%% file: Tables/reasoning_results.tex
\setlength{\aboverulesep}{0pt}\setlength{\belowrulesep}{0pt}\renewcommand{\arraystretch}{1.15}
\begin{tabular}{l|llcrrrrrr}
\toprule
Model & Bitwidth & Method & Bits (K/V) & AIME'24 & AIME'25 & GPQA-D & HumanEval+ & MBPP+ & Avg. \\
\midrule
\multirow{9}{*}{Ouro-1.4B-Thinking} &  & BF16 & 16/16 & 47.78$\pm$5.09 & 32.22$\pm$5.09 & 45.12$\pm$2.28 & 84.55$\pm$0.70 & 72.49$\pm$1.32 & 56.43$\pm$1.54 \\
\cmidrule(r){2-10}
 & \multirow{4}{*}{0.5 bit} & CQ & 0.5/0.5 & 1.11$\pm$1.92 & 2.22$\pm$1.92 & 11.78$\pm$1.05 & 27.85$\pm$0.70 & 35.27$\pm$2.31 & 15.65$\pm$0.76 \\
 &  & TaSQ & 0.508/0.5 & 4.44$\pm$1.92 & 5.56$\pm$1.92 & 12.12$\pm$2.31 & 35.16$\pm$2.14 & 40.12$\pm$2.72 & 19.48$\pm$0.99 \\
 &  & \cellcolor[HTML]{E8F1FB}\textbf{LoopVQ} & \cellcolor[HTML]{E8F1FB}0.5/0.5 & \cellcolor[HTML]{E8F1FB}38.89$\pm$5.09 & \cellcolor[HTML]{E8F1FB}31.11$\pm$3.85 & \cellcolor[HTML]{E8F1FB}40.57$\pm$1.27 & \cellcolor[HTML]{E8F1FB}75.81$\pm$1.27 & \cellcolor[HTML]{E8F1FB}67.64$\pm$1.19 & \cellcolor[HTML]{E8F1FB}50.80$\pm$1.35 \\
 &  & \cellcolor[HTML]{E8F1FB}TaSQ+\textbf{LoopVQ} & \cellcolor[HTML]{E8F1FB}0.508/0.5 & \cellcolor[HTML]{E8F1FB}\textbf{45.56$\pm$8.39} & \cellcolor[HTML]{E8F1FB}\textbf{34.44$\pm$5.09} & \cellcolor[HTML]{E8F1FB}\textbf{42.76$\pm$3.04} & \cellcolor[HTML]{E8F1FB}\textbf{81.50$\pm$1.53} & \cellcolor[HTML]{E8F1FB}\textbf{70.11$\pm$0.70} & \cellcolor[HTML]{E8F1FB}\textbf{54.87$\pm$2.08} \\
\cmidrule(r){2-10}
 & \multirow{4}{*}{0.25 bit} & CQ & 0.25/0.25 & 0.00$\pm$0.00 & 1.11$\pm$1.92 & 11.62$\pm$3.07 & 25.61$\pm$2.79 & 33.86$\pm$0.00 & 14.44$\pm$0.92 \\
 &  & TaSQ & 0.258/0.25 & 2.22$\pm$1.92 & 1.11$\pm$1.92 & 11.45$\pm$3.79 & 27.03$\pm$2.82 & 32.01$\pm$0.26 & 14.76$\pm$1.09 \\
 &  & \cellcolor[HTML]{E8F1FB}\textbf{LoopVQ} & \cellcolor[HTML]{E8F1FB}0.25/0.25 & \cellcolor[HTML]{E8F1FB}7.78$\pm$3.85 & \cellcolor[HTML]{E8F1FB}8.89$\pm$3.85 & \cellcolor[HTML]{E8F1FB}14.48$\pm$4.08 & \cellcolor[HTML]{E8F1FB}31.50$\pm$1.27 & \cellcolor[HTML]{E8F1FB}37.13$\pm$1.46 & \cellcolor[HTML]{E8F1FB}19.95$\pm$1.41 \\
 &  & \cellcolor[HTML]{E8F1FB}TaSQ+\textbf{LoopVQ} & \cellcolor[HTML]{E8F1FB}0.258/0.25 & \cellcolor[HTML]{E8F1FB}\textbf{36.67$\pm$3.33} & \cellcolor[HTML]{E8F1FB}\textbf{26.67$\pm$5.77} & \cellcolor[HTML]{E8F1FB}\textbf{36.53$\pm$4.58} & \cellcolor[HTML]{E8F1FB}\textbf{76.63$\pm$2.31} & \cellcolor[HTML]{E8F1FB}\textbf{63.84$\pm$0.85} & \cellcolor[HTML]{E8F1FB}\textbf{48.07$\pm$1.69} \\
\midrule
\multirow{9}{*}{Ouro-2.6B-Thinking} &  & BF16 & 16/16 & 57.78$\pm$11.71 & 51.11$\pm$5.09 & 54.04$\pm$3.31 & 86.18$\pm$0.93 & 75.49$\pm$0.81 & 64.92$\pm$2.65 \\
\cmidrule(r){2-10}
 & \multirow{4}{*}{0.5 bit} & CQ & 0.5/0.5 & 3.33$\pm$0.00 & 5.56$\pm$1.92 & 7.24$\pm$3.04 & 30.08$\pm$2.75 & 36.68$\pm$1.10 & 16.58$\pm$0.93 \\
 &  & TaSQ & 0.508/0.5 & 10.00$\pm$3.33 & 12.22$\pm$1.92 & 22.73$\pm$3.31 & 43.90$\pm$2.79 & 52.73$\pm$1.99 & 28.32$\pm$1.23 \\
 &  & \cellcolor[HTML]{E8F1FB}\textbf{LoopVQ} & \cellcolor[HTML]{E8F1FB}0.5/0.5 & \cellcolor[HTML]{E8F1FB}47.78$\pm$5.09 & \cellcolor[HTML]{E8F1FB}35.56$\pm$1.92 & \cellcolor[HTML]{E8F1FB}47.81$\pm$2.04 & \cellcolor[HTML]{E8F1FB}84.76$\pm$1.61 & \cellcolor[HTML]{E8F1FB}74.34$\pm$1.32 & \cellcolor[HTML]{E8F1FB}58.05$\pm$1.24 \\
 &  & \cellcolor[HTML]{E8F1FB}TaSQ+\textbf{LoopVQ} & \cellcolor[HTML]{E8F1FB}0.508/0.5 & \cellcolor[HTML]{E8F1FB}\textbf{61.11$\pm$3.85} & \cellcolor[HTML]{E8F1FB}\textbf{43.33$\pm$5.77} & \cellcolor[HTML]{E8F1FB}\textbf{51.01$\pm$4.63} & \cellcolor[HTML]{E8F1FB}\textbf{88.21$\pm$0.93} & \cellcolor[HTML]{E8F1FB}\textbf{75.49$\pm$1.50} & \cellcolor[HTML]{E8F1FB}\textbf{63.83$\pm$1.71} \\
\cmidrule(r){2-10}
 & \multirow{4}{*}{0.25 bit} & CQ & 0.25/0.25 & 2.22$\pm$1.92 & 3.33$\pm$3.33 & 8.25$\pm$1.54 & 30.89$\pm$1.96 & 38.80$\pm$2.90 & 16.70$\pm$1.09 \\
 &  & TaSQ & 0.258/0.25 & 1.11$\pm$1.92 & 5.56$\pm$1.92 & 10.94$\pm$0.77 & 33.33$\pm$0.93 & 39.42$\pm$0.26 & 18.07$\pm$0.60 \\
 &  & \cellcolor[HTML]{E8F1FB}\textbf{LoopVQ} & \cellcolor[HTML]{E8F1FB}0.25/0.25 & \cellcolor[HTML]{E8F1FB}10.00$\pm$3.33 & \cellcolor[HTML]{E8F1FB}5.56$\pm$1.92 & \cellcolor[HTML]{E8F1FB}13.13$\pm$3.31 & \cellcolor[HTML]{E8F1FB}43.50$\pm$2.14 & \cellcolor[HTML]{E8F1FB}48.50$\pm$1.25 & \cellcolor[HTML]{E8F1FB}24.14$\pm$1.13 \\
 &  & \cellcolor[HTML]{E8F1FB}TaSQ+\textbf{LoopVQ} & \cellcolor[HTML]{E8F1FB}0.258/0.25 & \cellcolor[HTML]{E8F1FB}\textbf{43.33$\pm$3.33} & \cellcolor[HTML]{E8F1FB}\textbf{34.44$\pm$5.09} & \cellcolor[HTML]{E8F1FB}\textbf{48.15$\pm$1.27} & \cellcolor[HTML]{E8F1FB}\textbf{82.72$\pm$0.35} & \cellcolor[HTML]{E8F1FB}\textbf{71.43$\pm$1.15} & \cellcolor[HTML]{E8F1FB}\textbf{56.02$\pm$1.27} \\
\bottomrule
\end{tabular}

%% file: Sections/6_Discussion.tex
\section{Discussion}
\label{sec:discussion}

\subsection{Ablation study}
\label{sec:ablation}

We ablate LoopVQ on Ouro-1.4B with GSM8K and HumanEval; all other settings follow
Section~\ref{sec:setup}.

\begin{table}[ht]
  \centering
  \small
  \setlength{\tabcolsep}{3.2pt}
  \begin{tabular}{lcccccc}
    \toprule
    & & & \multicolumn{2}{c}{GSM8K} & \multicolumn{2}{c}{HumanEval} \\
    \cmidrule(lr){4-5}\cmidrule(lr){6-7}
    Base & + Coupling & + FT & 0.5 & 0.25 & 0.5 & 0.25 \\
    \midrule
    \multirow{4}{*}{CQ}
      &            &            & 46.8 & 43.0 & 14.6 & 11.0 \\
      & \checkmark &            & 75.4 & 57.5 & 61.6 & 23.8 \\
      &            & \checkmark & 48.7 & 42.5 & 14.6 & 11.0 \\
      & \checkmark & \checkmark & \textbf{75.8} & \textbf{58.9} & \textbf{63.4} & \textbf{25.6} \\
    \midrule
    \multirow{4}{*}{TaSQ}
      &            &            & 57.3 & 44.9 & 24.4 & 12.2 \\
      & \checkmark &            & 77.5 & 68.7 & \textbf{70.1} & 50.6 \\
      &            & \checkmark & 59.7 & 47.1 & 31.7 & 14.0 \\
      & \checkmark & \checkmark & \textbf{77.6} & \textbf{71.0} & \textbf{70.1} & \textbf{55.5} \\
    \bottomrule
  \end{tabular}
  \caption{Loop-axis coupling and loop-aware fine-tuning (FT) applied to CQ and TaSQ on
  Ouro-1.4B at 0.5 and 0.25 bits per element. With both, CQ and TaSQ become LoopVQ and
  TaSQ+LoopVQ. BF16: 78.8 (GSM8K) and 73.2 (HumanEval).}
  \label{tab:abl_components}
\end{table}

\paragraph{Components.}
To isolate the contribution of each component, we apply loop-axis coupling and loop-aware
fine-tuning separately to CQ and TaSQ (Table~\ref{tab:abl_components}). Coupling accounts for most
of the gain at both bitwidths. Fine-tuning has little effect on GSM8K, whereas on HumanEval it tends
to improve accuracy, most clearly with TaSQ. Combining both components gives the best or tied-best
result in every setting. At 0.25 bit, coupling benefits TaSQ more than CQ.

\begin{figure}[ht]
  \centering
  \includegraphics[width=\columnwidth]{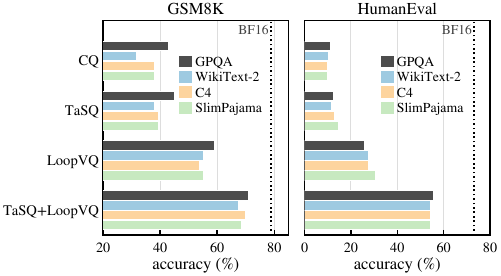}
  \caption{Accuracy of Ouro-1.4B at 0.25 bit with every method calibrated on a different
  corpus. GPQA is the calibration set of the main experiments.}
  \label{fig:calib_corpus}
\end{figure}

\paragraph{Calibration corpus.}
We next test whether LoopVQ depends on the choice of calibration data by recalibrating every
method on WikiText-2, C4, or SlimPajama (Figure~\ref{fig:calib_corpus}). The corpus noticeably
affects the baselines, with CQ varying by up to 11 points on GSM8K. TaSQ+LoopVQ is less
sensitive overall, varying by at most 4 points on GSM8K and 1 point on HumanEval, and remains the best
method on every corpus.

\begin{table}[ht]
  \centering
  \small
  \setlength{\tabcolsep}{4pt}
  \begin{tabular}{lcccccc}
    \toprule
    & \multicolumn{3}{c}{GSM8K} & \multicolumn{3}{c}{HumanEval} \\
    \cmidrule(lr){2-4}\cmidrule(lr){5-7}
    Sequences & 8 & 16 & 32 & 8 & 16 & 32 \\
    \midrule
    LoopVQ      & 57.5 & 58.9 & 60.8 & 22.6 & 25.6 & 28.0 \\
    TaSQ+LoopVQ & 69.3 & 71.0 & 71.9 & 49.4 & 55.5 & 59.8 \\
    \bottomrule
  \end{tabular}
  \caption{Accuracy of Ouro-1.4B at 0.25 bit versus the number of 2{,}048-token GPQA
  calibration sequences (16 in the main experiments).}
  \label{tab:abl_size}
\end{table}

\paragraph{Calibration-set size.}
We then vary the number of 2{,}048-token GPQA calibration sequences from 8 to 32
(Table~\ref{tab:abl_size}). Accuracy improves consistently as the set grows, most markedly on
HumanEval, and even the smallest set keeps both variants well ahead of the baselines. As
calibration takes only minutes (Section~\ref{sec:setup}), the set can be enlarged at little cost.

\subsection{Reasoning stability}
\label{sec:stability}

\begin{figure}[h]
  \centering
  \includegraphics[width=0.85\columnwidth]{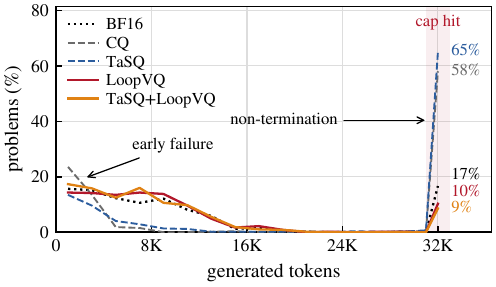}
  \vspace{-10pt}
  \caption{\textbf{Generation length on GPQA-Diamond} of Ouro-1.4B-Thinking under 0.5-bit KV cache
  quantization, with the share of problems that reach the 32K-token budget.}
  \label{fig:stability}
\end{figure}

When evaluating quantization methods, efficiency is usually measured as throughput or latency per generated token ignoring how many
tokens are needed to reach an answer. Compression errors can lengthen reasoning or keep it from terminating, significantly wasting computation and decoding time~\citep{liu2025hold, cheong2026kv}. On GPQA-Diamond, CQ and TaSQ exhaust the
32K-token budget on most problems and generate about twice as many tokens as BF16, whereas LoopVQ
and TaSQ+LoopVQ follow the length distribution of BF16 (Figure~\ref{fig:stability}). As a result, the throughput gains of LoopVQ
(Section~\ref{sec:efficiency}) carry over to the time needed to answer.
Appendix~\ref{app:stability} extends this analysis, including early termination, to both models and bitwidths.

%% file: Sections/7_Related_Work.tex
\section{Related work}
\label{sec:relatedwork}

\paragraph{Quantization of LLMs.}
Weight quantization methods for LLMs include scalar quantization with error compensation or
activation-aware scaling~\cite{frantar2022gptq,lin2024awq}, joint quantization of weights and
activations~\cite{xiao2023smoothquant,ashkboos2024quarot}, and vector quantization with
codebooks~\cite{tseng2024quip,egiazarian2024extreme}. For the KV cache, scalar methods quantize keys
and values per channel or per token~\cite{liu2024kivi,hooper2024kvquant,ashkboos2024quarot}, and
vector quantization methods code groups of channels jointly~\cite{zhang2024kv,cheong2026tailoring,son2026nsnquant}.
These methods target non-looped models, where each layer writes one cache, and quantize the caches
of different layers independently.

\paragraph{Quantization of looped transformers.}
The recurrence of looped transformers has motivated quantization methods tailored to it. LoopQ~\cite{fang2026loopq}, LoopQuant~\cite{zhao2026loopquant}, and
Loopy~\cite{li2026loopy} quantize the weights and activations of the shared block, accounting for
the distribution shift and error accumulation across loop steps. For the KV cache, which dominates
the memory of a looped model (Figure~\ref{fig:motivation-overview}a), FlashLoop~\cite{yang2026flashloop} and
ResidualQuant~\cite{kim2026residualquant} apply scalar quantization at 2 to 4 bits per
element. LoopVQ instead applies vector quantization along the loop axis to a pretrained
model, so that one index codes several loop steps, and compresses the KV cache below one bit per
element.

%% file: Sections/8_Conclusion.tex
\section{Conclusion}
\label{sec:conclusion}

We presented LoopVQ, a KV cache quantization method for looped transformers. LoopVQ groups
channels of multiple loop steps into a sub-vector that shares one index, and fine-tunes the
codebooks to match the output of the full-precision model, accounting for the different
quantization sensitivities of the loop steps. We showed that, at 0.5 and 0.25 bits
per element, LoopVQ outperforms existing vector quantization methods on general, reasoning, and
long-context benchmarks, and that our SGLang implementation reduces the decode latency and raises
the serving throughput over the BF16 baseline. The appendix
provides further analysis of the fine-tuning objective, implementation and kernel optimization details, and additional experiment results.

%% file: Sections/Limitations.tex
Our efficiency evaluation compares LoopVQ with BF16 serving in SGLang and does not include the
baseline quantization methods. The serving efficiency of each method depends on kernels optimized
for its own cache format, and it is difficult to determine a comparable level of optimization for
every method. To avoid a comparison biased toward the method we optimized most, we compare against
BF16 serving, which SGLang already supports with highly optimized kernels.

We also do not compare accuracy with FlashLoop~\cite{yang2026flashloop} and
ResidualQuant~\cite{kim2026residualquant}, the existing KV cache quantization methods for looped
transformers. Both apply scalar quantization, which stores at least one bit per element and cannot
reach the sub-1-bit regime that LoopVQ targets. They are designed for 2 to 4 bits per element and
do not specify a configuration for lower bitwidths, so they cannot be compared with LoopVQ at
matched bitwidths.

%% file: Sections/Appendix.tex
\section{Analysis of loop-aware fine-tuning}
\label{app:ft}

This appendix supports two statements of Section~\ref{sec:ft}. First, the shortcomings of
Fisher-weighted $k$-means cannot be removed by changing its weights, which is why we optimize the
output objective~\eqref{eq:ft} directly (Appendix~\ref{app:ft-fisher}). Second, although code
assignment is discrete, the codebook entries receive an exact gradient and can be trained by
ordinary backpropagation (Appendix~\ref{app:ft-grad}).

Throughout, $s$ is a calibration sequence, $C$ the cache of its unquantized forward pass, and
$\delta=\hat C(\mathcal{B})-C$ the quantization error, viewed as one vector over all tokens, loop
steps, layers, heads, and channels.

\subsection{Limitation of weighted $k$-means}
\label{app:ft-fisher}

A natural alternative to fine-tuning is to improve the weighted $k$-means calibration itself~\eqref{eq:wkmeans}. We show that
no choice of weights can represent the fine-tuning objective~\eqref{eq:ft}.

\paragraph{The objective couples coded vectors.}
For small $\delta$, the KL divergence is quadratic in the error,
\begin{equation}
  \mathcal{J}(\mathcal{B})
  \approx \tfrac12\,\mathbb{E}_s\bigl[\delta^{\top}F\delta\bigr],
  \qquad F=\tfrac1N\textstyle\sum_jF_j ,
  \label{eq:app-quad}
\end{equation}
where $F_j$ is the Fisher information of $q_j$ with respect to the cache. Splitting $\delta$ into the
errors $\delta_n=e_{a(z_n)}-z_n$ of the individual coded vectors~\eqref{eq:coupled},
\begin{equation}
  \begin{aligned}
  \delta^{\top}F\delta
  ={}& \textstyle\sum_{n}\delta_n^{\top}F_{nn}\,\delta_n \\
     & + \textstyle\sum_{n\neq n'}\delta_n^{\top}F_{nn'}\,\delta_{n'} ,
  \end{aligned}
  \label{eq:app-blocks}
\end{equation}
where $F_{nn'}$ is the block of $F$ that pairs vectors $n$ and $n'$. The cross terms
$F_{nn'}$ connect, among others, the codes of a token at different loop steps and layers: by the
recurrence~\eqref{eq:loop}, the error of an early loop step propagates to the later loop steps and interacts with
their errors.

\paragraph{A weighted $k$-means objective cannot express the cross terms.}
Any weighted $k$-means objective, Fisher-weighted $k$-means included, is a sum of terms that each
depend on a single coded vector, $\sum_n\delta_n^{\top}W_n\delta_n$; Fisher-weighted $k$-means sets
$W_n=\omega_nI$. Since each term involves only one vector, $k$-means can assign every vector to its
nearest entry independently of the others and update each entry from the vectors assigned to it.
For the same reason, such an objective can at best match the diagonal blocks $F_{nn}$
of~\eqref{eq:app-blocks} and always drops the cross terms. A better $W_n$ could account for the unequal sensitivity of the loop
steps within a coded vector (Section~\ref{sec:motivation}), but not for the interaction between
coded vectors. Fine-tuning minimizes~\eqref{eq:ft} itself, which keeps both and, unlike any
quadratic surrogate, also holds for large errors.

\subsection{Gradient with respect to the codebook entries}
\label{app:ft-grad}

Vector quantization selects entries by an $\arg\min$, which has no useful derivative. We show that
the codebook entries nevertheless receive an exact gradient, because the selection does not need
to be differentiated.

The reconstruction $\hat C(\mathcal{B})$ depends on $\mathcal{B}$ through the assignments
$a(z_j)$ and the decoding~\eqref{eq:decode}. The coupled vectors $z_j$ are formed from the
unquantized cache $C$ and do not depend on $\mathcal{B}$. Each assignment $a(z_j)$ is constant as
long as the nearest entry of $z_j$ does not change, i.e., everywhere except on the measure-zero
set of codebooks for which $z_j$ is equidistant from two entries. Off that set,
$\partial a(z_j)/\partial\mathcal{B}=0$, and~\eqref{eq:decode}, which copies coordinate $iT_c+m$ of
the selected entry into the reconstruction of loop step $\beta T_c+m+1$, is linear in that entry. For an
entry $e_r$ of $\mathcal{B}_{\beta,\ell,h,g}$, the chain rule gives
\begin{equation}
  \Bigl[\frac{\partial\mathcal{J}}{\partial e_r}\Bigr]_{iT_c+m}
  =\sum_{j:\,a(z_j)=r}
  \frac{\partial\mathcal{J}}{\partial \hat x^{(\beta T_c+m+1,\ell)}_{j,h,gc+i}}
  \label{eq:ft-grad}
\end{equation}
for $i<c$ and $m<T_c$, where the sum runs over the calibration tokens assigned to $e_r$. Each entry accumulates the
gradient of every token assigned to it, at each of the $T_c$ loop steps it covers. Because the
coupled vectors $z_j$ do not depend on $\mathcal{B}$, no straight-through estimator is needed.

In practice, the backward pass runs in two stages: through the network down to the reconstructed
cache, and then from the cache to the entries via~\eqref{eq:ft-grad}, in chunks of tokens. This
yields the same gradient as a single backward pass through one graph while bounding the memory of
the second stage. After each update, the assignments are recomputed with the current codebooks.

\section{Orthogonal application with TaSQ}
\label{app:tasq}

TaSQ~\cite{cheong2026tailoring} improves key quantization along the channel axis: it transforms the channels of
each key so that those sharing a code are correlated and the reconstruction error reflects the
attention logits. LoopVQ is orthogonal
to it, as it shares codes along the loop axis. TaSQ+LoopVQ applies TaSQ's transform before
loop-axis coupling, so that each code groups correlated channels across correlated loop steps.

\paragraph{Formulation.}
TaSQ maps the pre-RoPE key of loop step $t$ in block $\beta$, layer $\ell$, and head $h$ to
\begin{equation}
  u^{(t,\ell)}_{j,h} = P_{\beta,\ell,h}\,
    \frac{w^{(t,\ell)}_{h}\odot k^{(t,\ell)}_{j,h}}{\sigma^{(t,\ell)}_j},
  \label{eq:tasq}
\end{equation}
where $w^{(t,\ell)}_h$ weights each channel by the second moment $M$ of the post-RoPE queries that
read this cache, $w_i^2\propto M_{ii}$, $\sigma^{(t,\ell)}_j$ is the root mean square of
$w\odot k$ over all heads and channels of token $j$, and $P_{\beta,\ell,h}$ is a channel
permutation. TaSQ+LoopVQ builds the coupled key vectors~\eqref{eq:coupled} from $u$ instead of $k$,
\begin{equation}
  \bigl[z^{(\beta,\ell,h,g)}_j\bigr]_{iT_c+m} = u^{(\beta T_c+m+1,\,\ell)}_{j,h,\,gc+i},
  \quad i<c,\; m<T_c ,
  \label{eq:tasq-coupled}
\end{equation}
so that each code spans $c$ channels chosen by $P$ along the channel axis and $T_c$ loop steps along
the loop axis.

\paragraph{Grouping under loop coupling.}
In TaSQ, $P$ groups channels with high covariance into the same code, keeping each RoPE pair
$\{i,i+D/2\}$ together so that RoPE can be applied after decoding. Under~\eqref{eq:tasq-coupled},
a code contains the same channels at all $T_c$ loop steps, so the grouping must be shared by the loop steps
of a block. We compute $P_{\beta,\ell,h}$ from the covariance of the transformed keys of
the whole block,
\begin{equation}
  \Sigma_{\beta,\ell,h}
  = \mathrm{Cov}_j\bigl[\tilde k^{(\beta T_c+1,\ell)}_{j,h};\dots;
    \tilde k^{(\beta T_c+T_c,\ell)}_{j,h}\bigr],
  \label{eq:tasq-cov}
\end{equation}
a $T_cD\times T_cD$ matrix over the stacked loop steps with $\tilde k=w\odot k/\sigma$, and group RoPE pairs taken at all $T_c$ loop steps ($2T_c$ coordinates)
until a group holds $G=cT_c$ coordinates. This grouping accounts for the correlation between
channels both within and across loop steps.

\paragraph{Calibration and bitwidth.}
The Fisher weights of~\eqref{eq:wkmeans} are computed with respect to $u$. During
fine-tuning (Section~\ref{sec:ft}), $w$ and $P$ are fixed and $\sigma$ is recomputed from the keys.
Storing $\sigma$ in fp16 costs $16/(HD)$ bits per key element, about $0.008$ on Ouro and $0.003$ on
Huginn, which gives the key bitwidths of TaSQ and TaSQ+LoopVQ in Table~\ref{tab:general_results}.

\section{Optimizing pre-RoPE vector quantization serving efficiency}
\label{app:kernel}

LoopVQ, like CQ and TaSQ, quantizes keys before the rotary embedding (RoPE)
(Section~\ref{sec:prelim}), since RoPE rotates the same key differently at every position and
would spread the keys of one channel group over many more codebook entries. The decode attention
kernel then has to apply RoPE to every reconstructed key (Algorithm~\ref{alg:vq_attention}), which requires the
cosine and sine of $D/2$ frequencies at the position of each cached token. Evaluating them in the
kernel adds transcendental operations for every token and frequency, so the kernel reads them from
a table precomputed for all positions, one row of $D$ values per position.

With one request per thread block, every block reads the table row of each token it attends to,
and the blocks of different requests read the same rows again. At large batch sizes these reads
account for about 40\% of the kernel time. Since a token's RoPE position depends only on its
index in the sequence, the rows needed by two requests coincide wherever both requests have a
token. The kernel assigns two requests to each thread block and cuts both into splits at
the same boundaries, sized by the longer request, so that one load of a table row serves the
tokens of both requests at the same index. The two requests also share the codebook slices
$\mathcal{B}^k_t$ and $\mathcal{B}^v_t$ staged in shared memory, which halves the codebook loads per
request. The shared-memory buffers of each request are halved accordingly, so a block uses the same
shared memory as with one request. Each lane further handles the two channel groups that share the
same frequencies, so that it loads their cosines and sines with single vector loads. At batch size 1
the kernel processes one request per block.

On an RTX~6000 Ada GPU with Ouro-1.4B-Thinking, this kernel reduces the decode
attention time at batch size 64 and a 6K-token context by 37\% and 48\% at 0.5 and 0.25 bit
compared with one request per block, and
raises the end-to-end throughput at batch size 64 with 2K-token prompts and 1K-token outputs by
46\% and 63\%.

\section{Additional experiment results}
\subsection{Efficiency results on an A100 (40\,GB) GPU}
\label{app:efficiency}

\begin{figure}[t]
  \centering
  \includegraphics[width=\columnwidth]{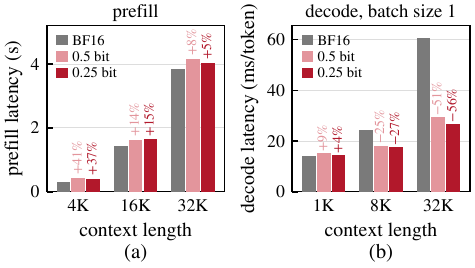}
  \caption{Latency of LoopVQ at 0.5 and 0.25 bit on Ouro-1.4B-Thinking with one A100 (40\,GB) GPU.
  (a)~Prefill latency and (b)~decode latency per token at batch size 1, each with the change
  relative to BF16.}
  \label{fig:latency_a100}
\end{figure}

\begin{figure*}[t]
  \centering
  \includegraphics[width=\textwidth]{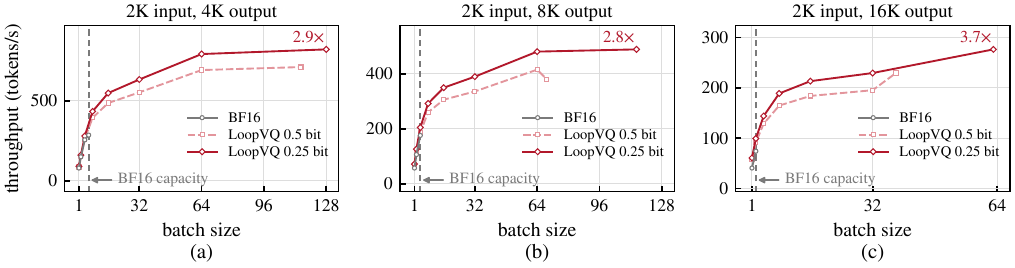}
  \caption{End-to-end throughput of LoopVQ on Ouro-1.4B-Thinking with one A100 (40\,GB) GPU, versus
  batch size up to the largest batch that fits in memory.}
  \label{fig:throughput_a100}
\end{figure*}

We repeat the measurements of Section~\ref{sec:efficiency} on one NVIDIA A100 (40\,GB) GPU with the
same settings. The results follow the same trends as on the RTX~6000 Ada. The prefill overhead
of LoopVQ decreases with the context length, and the reduction in decode latency grows with it,
reaching about half of the BF16 latency at 32K tokens (Figure~\ref{fig:latency_a100}). As BF16
fits only a few long requests in memory, the peak throughput of LoopVQ exceeds that of BF16 by
$2.8\times$ to $3.7\times$ (Figure~\ref{fig:throughput_a100}). On the A100, the $k$-means calibration and the fine-tuning take 5.5 and 2.0
minutes for Ouro-1.4B and 11.2 and 4.6 minutes for Ouro-2.6B.

\subsection{Efficiency results on Ouro-2.6B-Thinking}
\label{app:efficiency_2p6b}

\begin{figure}[t]
  \centering
  \includegraphics[width=\columnwidth]{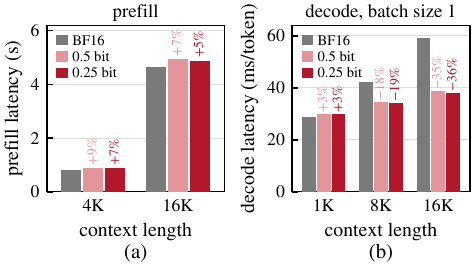}
  \caption{Latency of LoopVQ at 0.5 and 0.25 bit on Ouro-2.6B-Thinking with one RTX~6000 Ada
  GPU. (a)~Prefill latency and (b)~decode latency per token at batch size 1, each with the change
  relative to BF16.}
  \label{fig:latency_2p6b}
\end{figure}

\begin{figure*}[t]
  \centering
  \includegraphics[width=\textwidth]{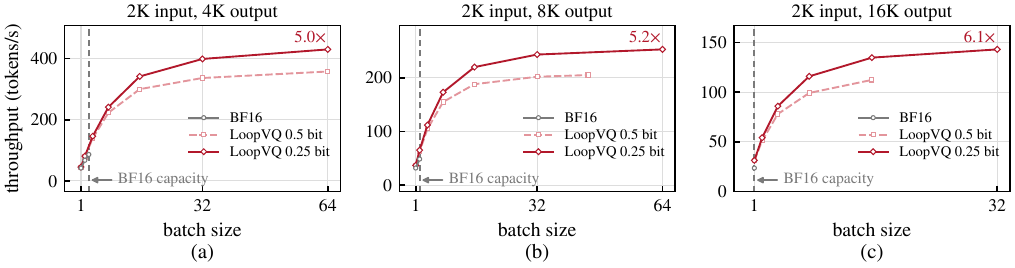}
  \caption{End-to-end throughput of LoopVQ on Ouro-2.6B-Thinking with one RTX~6000 Ada GPU, versus
  batch size up to the largest batch that fits in memory.}
  \label{fig:throughput_2p6b}
\end{figure*}

We repeat the measurements of Section~\ref{sec:efficiency} on Ouro-2.6B-Thinking with the same GPU
and settings. Prefill with LoopVQ is 5 to 9\% slower than with BF16, and the decode latency at
16K tokens is about 35\% lower (Figure~\ref{fig:latency_2p6b}). As Ouro-2.6B has twice as many
layers as Ouro-1.4B, BF16 fits only one to three long requests in memory, and the peak throughput
of LoopVQ exceeds that of BF16 by $5.0\times$ to $6.1\times$ (Figure~\ref{fig:throughput_2p6b}).

\begin{figure*}[t]
  \centering
  \includegraphics[width=\textwidth]{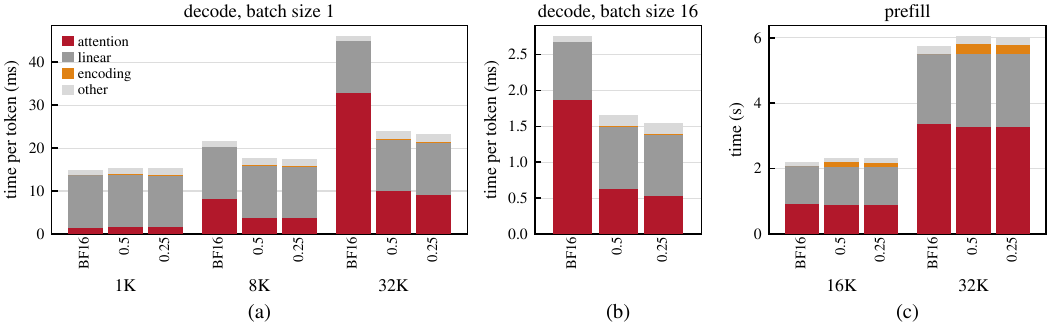}
  \caption{GPU time of LoopVQ and BF16 on Ouro-1.4B-Thinking with one RTX~6000 Ada GPU, by
  kernel type. (a)~Decode time per token at batch size 1. (b)~Decode time per generated token at
  batch size 16 with 2K-token prompts. (c)~Prefill time. Attention includes the bf16 window, the
  LoopVQ kernel, and the merge of their splits.}
  \label{fig:profile}
\end{figure*}

\subsection{Time profile breakdown}
\label{app:profile}

Figure~\ref{fig:profile} breaks down the GPU time by kernel type, measured with the PyTorch
profiler under the settings of Section~\ref{sec:efficiency}. During decoding, the linear layers
take the same time in all methods, and the difference comes from attention. With BF16, attention
grows with the context length and accounts for most of the decode time at 32K tokens, whereas
with LoopVQ it stays below the time of the linear layers. At batch size 16, attention takes
1.87\,ms per generated token with BF16 and 0.63 and 0.54\,ms with LoopVQ at 0.5 and 0.25 bit.
Encoding takes under 1\% of the decode time and 5 to 7\% of the prefill time.

\begin{figure*}[t]
  \centering
  \includegraphics[width=\textwidth]{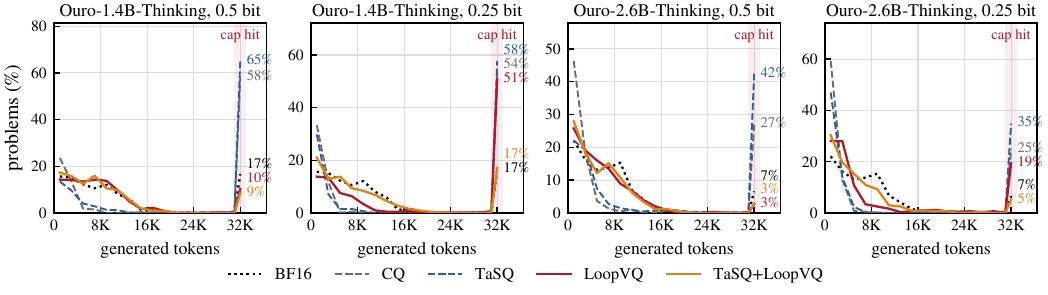}
  \caption{Generation length on GPQA-Diamond of Ouro-1.4B-Thinking and Ouro-2.6B-Thinking under
  0.5- and 0.25-bit KV cache quantization, with the share of problems that reach the 32K-token
  budget.}
  \label{fig:stability_all}
\end{figure*}

\subsection{Reasoning stability across models and bitwidths}
\label{app:stability}

Figure~\ref{fig:stability_all} extends Figure~\ref{fig:stability} to both Thinking models and both
bitwidths. CQ and TaSQ exhaust the generation budget on a large share of problems in every setting.
TaSQ+LoopVQ follows the length distribution of BF16 in all four settings, and LoopVQ does so at
0.5 bit.

Quantization errors can also end the reasoning early. With BF16, 16\% and 22\% of the generations
of Ouro-1.4B-Thinking and Ouro-2.6B-Thinking end within 2K tokens. At 0.25 bit, this share rises to
33\% and 59\% with CQ and to 30\% and 47\% with TaSQ, whereas it stays at or below 30\% with
LoopVQ and TaSQ+LoopVQ in every setting.

\subsection{LongBench results}
\label{app:longbench}

\begin{table*}[t!]
\centering
\caption{LongBench performance of Ouro-1.4B and Ouro-2.6B under low-bit KV cache quantization on the
16 English tasks, with the official metric of each task. All results use greedy decoding. Bold
denotes the best quantized result; higher is better.}
\label{tab:longbench_results}
\resizebox{\linewidth}{!}{\input{Tables/longbench_results}}
\end{table*}

We also evaluate the base models on the 16 English tasks of LongBench~\citep{bai2024longbench},
using the official prompts and metrics, greedy decoding, and prompts truncated in the middle to
64K tokens. As in the other long-context experiments, 64 sink and 256 recent tokens are kept in
bf16. CQ and TaSQ fall 7 to 13 points below the BF16 average (Table~\ref{tab:longbench_results}).
LoopVQ outperforms both baselines on average in every setting, and TaSQ+LoopVQ comes within
1.5 points of BF16 at 0.5 bit and achieves the highest average at both bitwidths.

%% file: Tables/longbench_results.tex
\setlength{\aboverulesep}{0pt}\setlength{\belowrulesep}{0pt}\renewcommand{\arraystretch}{1.15}
\begin{tabular}{l|llrrrrrrrrrrrrrrrrr}
\toprule
\multirow{2}{*}{Model} & \multirow{2}{*}{Bitwidth} & \multirow{2}{*}{Method} & \multicolumn{3}{c}{Single-doc QA} & \multicolumn{3}{c}{Multi-doc QA} & \multicolumn{3}{c}{Summarization} & \multicolumn{3}{c}{Few-shot} & \multicolumn{2}{c}{Synthetic} & \multicolumn{2}{c}{Code} & \multirow{2}{*}{Avg.} \\
\cmidrule(lr){4-6}\cmidrule(lr){7-9}\cmidrule(lr){10-12}\cmidrule(lr){13-15}\cmidrule(lr){16-17}\cmidrule(lr){18-19}
 & & & \rotatebox{60}{NrtvQA} & \rotatebox{60}{Qasper} & \rotatebox{60}{MF-en} & \rotatebox{60}{HotpotQA} & \rotatebox{60}{2WikiMQA} & \rotatebox{60}{Musique} & \rotatebox{60}{GovReport} & \rotatebox{60}{QMSum} & \rotatebox{60}{MultiNews} & \rotatebox{60}{TREC} & \rotatebox{60}{TriviaQA} & \rotatebox{60}{SAMSum} & \rotatebox{60}{PCount} & \rotatebox{60}{PRe} & \rotatebox{60}{Lcc} & \rotatebox{60}{RB-P} & \\
\midrule
\multirow{9}{*}{Ouro-1.4B} &  & BF16 & 7.81 & 23.56 & 34.75 & 33.17 & 19.75 & 13.15 & 26.94 & 20.52 & 23.08 & 64.50 & 88.57 & 36.18 & 1.50 & 32.50 & 19.99 & 28.72 & 29.67 \\
\cmidrule(r){2-20}
 & \multirow{4}{*}{0.5 bit} & CQ & 4.93 & 10.38 & 22.10 & 20.18 & 16.35 & 8.08 & 16.88 & 17.44 & 15.72 & 44.00 & 79.12 & 33.07 & 1.50 & 4.50 & 19.68 & 27.95 & 21.37 \\
 &  & TaSQ & 7.02 & 13.39 & 25.05 & 24.90 & 17.46 & 9.22 & 18.38 & 17.88 & 16.58 & 47.00 & 81.79 & 34.07 & 1.00 & 4.50 & 21.85 & 27.87 & 23.00 \\
 &  & \cellcolor[HTML]{E8F1FB}\textbf{LoopVQ} & \cellcolor[HTML]{E8F1FB}7.31 & \cellcolor[HTML]{E8F1FB}19.29 & \cellcolor[HTML]{E8F1FB}30.71 & \cellcolor[HTML]{E8F1FB}32.63 & \cellcolor[HTML]{E8F1FB}\textbf{19.23} & \cellcolor[HTML]{E8F1FB}\textbf{12.82} & \cellcolor[HTML]{E8F1FB}23.73 & \cellcolor[HTML]{E8F1FB}18.94 & \cellcolor[HTML]{E8F1FB}21.06 & \cellcolor[HTML]{E8F1FB}58.50 & \cellcolor[HTML]{E8F1FB}87.80 & \cellcolor[HTML]{E8F1FB}35.61 & \cellcolor[HTML]{E8F1FB}\textbf{2.00} & \cellcolor[HTML]{E8F1FB}16.00 & \cellcolor[HTML]{E8F1FB}22.09 & \cellcolor[HTML]{E8F1FB}29.60 & \cellcolor[HTML]{E8F1FB}27.33 \\
 &  & \cellcolor[HTML]{E8F1FB}TaSQ+\textbf{LoopVQ} & \cellcolor[HTML]{E8F1FB}\textbf{8.11} & \cellcolor[HTML]{E8F1FB}\textbf{22.84} & \cellcolor[HTML]{E8F1FB}\textbf{33.34} & \cellcolor[HTML]{E8F1FB}\textbf{34.56} & \cellcolor[HTML]{E8F1FB}18.89 & \cellcolor[HTML]{E8F1FB}12.60 & \cellcolor[HTML]{E8F1FB}\textbf{26.07} & \cellcolor[HTML]{E8F1FB}\textbf{19.68} & \cellcolor[HTML]{E8F1FB}\textbf{22.15} & \cellcolor[HTML]{E8F1FB}\textbf{64.00} & \cellcolor[HTML]{E8F1FB}\textbf{89.03} & \cellcolor[HTML]{E8F1FB}\textbf{36.09} & \cellcolor[HTML]{E8F1FB}1.50 & \cellcolor[HTML]{E8F1FB}\textbf{23.00} & \cellcolor[HTML]{E8F1FB}\textbf{22.16} & \cellcolor[HTML]{E8F1FB}\textbf{30.55} & \cellcolor[HTML]{E8F1FB}\textbf{29.04} \\
\cmidrule(r){2-20}
 & \multirow{4}{*}{0.25 bit} & CQ & 4.29 & 9.22 & 21.13 & 20.51 & 16.83 & 7.17 & 15.79 & 17.11 & 14.96 & 44.50 & 78.44 & 33.01 & 1.00 & 4.50 & 20.55 & 28.45 & 21.09 \\
 &  & TaSQ & 5.69 & 10.46 & 21.65 & 20.73 & 16.56 & 6.76 & 16.50 & 17.22 & 15.69 & 45.00 & 79.20 & 31.98 & 1.00 & 4.50 & 20.54 & 27.69 & 21.32 \\
 &  & \cellcolor[HTML]{E8F1FB}\textbf{LoopVQ} & \cellcolor[HTML]{E8F1FB}5.81 & \cellcolor[HTML]{E8F1FB}13.18 & \cellcolor[HTML]{E8F1FB}22.98 & \cellcolor[HTML]{E8F1FB}24.18 & \cellcolor[HTML]{E8F1FB}17.05 & \cellcolor[HTML]{E8F1FB}9.30 & \cellcolor[HTML]{E8F1FB}19.16 & \cellcolor[HTML]{E8F1FB}17.79 & \cellcolor[HTML]{E8F1FB}17.00 & \cellcolor[HTML]{E8F1FB}45.00 & \cellcolor[HTML]{E8F1FB}82.50 & \cellcolor[HTML]{E8F1FB}34.21 & \cellcolor[HTML]{E8F1FB}\textbf{1.50} & \cellcolor[HTML]{E8F1FB}4.50 & \cellcolor[HTML]{E8F1FB}21.69 & \cellcolor[HTML]{E8F1FB}28.97 & \cellcolor[HTML]{E8F1FB}22.80 \\
 &  & \cellcolor[HTML]{E8F1FB}TaSQ+\textbf{LoopVQ} & \cellcolor[HTML]{E8F1FB}\textbf{6.40} & \cellcolor[HTML]{E8F1FB}\textbf{19.61} & \cellcolor[HTML]{E8F1FB}\textbf{30.48} & \cellcolor[HTML]{E8F1FB}\textbf{32.57} & \cellcolor[HTML]{E8F1FB}\textbf{18.71} & \cellcolor[HTML]{E8F1FB}\textbf{12.67} & \cellcolor[HTML]{E8F1FB}\textbf{22.88} & \cellcolor[HTML]{E8F1FB}\textbf{18.36} & \cellcolor[HTML]{E8F1FB}\textbf{19.77} & \cellcolor[HTML]{E8F1FB}\textbf{56.50} & \cellcolor[HTML]{E8F1FB}\textbf{87.09} & \cellcolor[HTML]{E8F1FB}\textbf{34.39} & \cellcolor[HTML]{E8F1FB}1.00 & \cellcolor[HTML]{E8F1FB}\textbf{6.50} & \cellcolor[HTML]{E8F1FB}\textbf{22.05} & \cellcolor[HTML]{E8F1FB}\textbf{29.69} & \cellcolor[HTML]{E8F1FB}\textbf{26.17} \\
\midrule
\multirow{9}{*}{Ouro-2.6B} &  & BF16 & 9.90 & 31.54 & 47.80 & 42.38 & 39.16 & 28.00 & 28.08 & 23.53 & 25.73 & 72.00 & 88.76 & 13.50 & 0.00 & 70.00 & 36.95 & 37.46 & 37.18 \\
\cmidrule(r){2-20}
 & \multirow{4}{*}{0.5 bit} & CQ & 7.49 & 15.86 & 31.73 & 26.29 & 31.49 & 13.42 & 17.58 & 19.18 & 16.70 & 48.50 & 81.28 & 12.04 & 0.00 & 7.00 & 38.91 & 33.60 & 25.07 \\
 &  & TaSQ & 8.41 & 17.82 & 35.24 & 32.51 & 33.24 & 20.24 & 19.03 & 19.17 & 17.74 & 52.00 & 85.74 & 11.25 & 0.00 & 9.00 & \textbf{40.47} & 35.81 & 27.35 \\
 &  & \cellcolor[HTML]{E8F1FB}\textbf{LoopVQ} & \cellcolor[HTML]{E8F1FB}\textbf{10.21} & \cellcolor[HTML]{E8F1FB}25.53 & \cellcolor[HTML]{E8F1FB}45.30 & \cellcolor[HTML]{E8F1FB}41.84 & \cellcolor[HTML]{E8F1FB}35.03 & \cellcolor[HTML]{E8F1FB}23.20 & \cellcolor[HTML]{E8F1FB}22.61 & \cellcolor[HTML]{E8F1FB}21.32 & \cellcolor[HTML]{E8F1FB}22.39 & \cellcolor[HTML]{E8F1FB}62.00 & \cellcolor[HTML]{E8F1FB}88.93 & \cellcolor[HTML]{E8F1FB}\textbf{13.21} & \cellcolor[HTML]{E8F1FB}0.00 & \cellcolor[HTML]{E8F1FB}27.00 & \cellcolor[HTML]{E8F1FB}40.38 & \cellcolor[HTML]{E8F1FB}39.22 & \cellcolor[HTML]{E8F1FB}32.39 \\
 &  & \cellcolor[HTML]{E8F1FB}TaSQ+\textbf{LoopVQ} & \cellcolor[HTML]{E8F1FB}9.83 & \cellcolor[HTML]{E8F1FB}\textbf{29.41} & \cellcolor[HTML]{E8F1FB}\textbf{48.46} & \cellcolor[HTML]{E8F1FB}\textbf{43.18} & \cellcolor[HTML]{E8F1FB}\textbf{38.93} & \cellcolor[HTML]{E8F1FB}\textbf{26.11} & \cellcolor[HTML]{E8F1FB}\textbf{25.80} & \cellcolor[HTML]{E8F1FB}\textbf{22.38} & \cellcolor[HTML]{E8F1FB}\textbf{24.39} & \cellcolor[HTML]{E8F1FB}\textbf{72.00} & \cellcolor[HTML]{E8F1FB}\textbf{89.01} & \cellcolor[HTML]{E8F1FB}12.87 & \cellcolor[HTML]{E8F1FB}0.00 & \cellcolor[HTML]{E8F1FB}\textbf{49.00} & \cellcolor[HTML]{E8F1FB}39.73 & \cellcolor[HTML]{E8F1FB}\textbf{39.93} & \cellcolor[HTML]{E8F1FB}\textbf{35.69} \\
\cmidrule(r){2-20}
 & \multirow{4}{*}{0.25 bit} & CQ & 7.48 & 15.22 & 31.20 & 25.60 & 30.73 & 12.91 & 16.84 & 18.84 & 15.17 & 47.00 & 81.12 & 12.07 & 0.00 & 7.50 & 38.51 & 33.85 & 24.63 \\
 &  & TaSQ & 6.93 & 15.19 & 31.54 & 26.27 & 31.36 & 14.86 & 17.13 & 18.66 & 15.96 & 47.50 & 82.37 & 11.04 & 0.00 & 7.50 & 39.37 & 34.48 & 25.01 \\
 &  & \cellcolor[HTML]{E8F1FB}\textbf{LoopVQ} & \cellcolor[HTML]{E8F1FB}\textbf{8.93} & \cellcolor[HTML]{E8F1FB}18.13 & \cellcolor[HTML]{E8F1FB}35.29 & \cellcolor[HTML]{E8F1FB}30.69 & \cellcolor[HTML]{E8F1FB}32.35 & \cellcolor[HTML]{E8F1FB}16.59 & \cellcolor[HTML]{E8F1FB}19.27 & \cellcolor[HTML]{E8F1FB}19.49 & \cellcolor[HTML]{E8F1FB}18.64 & \cellcolor[HTML]{E8F1FB}50.50 & \cellcolor[HTML]{E8F1FB}86.14 & \cellcolor[HTML]{E8F1FB}\textbf{12.56} & \cellcolor[HTML]{E8F1FB}0.00 & \cellcolor[HTML]{E8F1FB}8.00 & \cellcolor[HTML]{E8F1FB}38.81 & \cellcolor[HTML]{E8F1FB}36.40 & \cellcolor[HTML]{E8F1FB}26.99 \\
 &  & \cellcolor[HTML]{E8F1FB}TaSQ+\textbf{LoopVQ} & \cellcolor[HTML]{E8F1FB}8.90 & \cellcolor[HTML]{E8F1FB}\textbf{23.42} & \cellcolor[HTML]{E8F1FB}\textbf{43.60} & \cellcolor[HTML]{E8F1FB}\textbf{41.73} & \cellcolor[HTML]{E8F1FB}\textbf{34.72} & \cellcolor[HTML]{E8F1FB}\textbf{25.40} & \cellcolor[HTML]{E8F1FB}\textbf{21.62} & \cellcolor[HTML]{E8F1FB}\textbf{20.29} & \cellcolor[HTML]{E8F1FB}\textbf{20.93} & \cellcolor[HTML]{E8F1FB}\textbf{66.50} & \cellcolor[HTML]{E8F1FB}\textbf{88.51} & \cellcolor[HTML]{E8F1FB}12.24 & \cellcolor[HTML]{E8F1FB}0.00 & \cellcolor[HTML]{E8F1FB}\textbf{28.50} & \cellcolor[HTML]{E8F1FB}\textbf{40.64} & \cellcolor[HTML]{E8F1FB}\textbf{38.56} & \cellcolor[HTML]{E8F1FB}\textbf{32.22} \\
\bottomrule
\end{tabular}